\documentclass[sn-mathphys-num, iicol]{sn-jnl}

\usepackage{graphicx}%
\usepackage{multirow}%
\usepackage{amsmath,amssymb,amsfonts}%
\usepackage{amsthm}%
\usepackage{mathrsfs}%
\usepackage[title]{appendix}%
\usepackage{xcolor}%
\usepackage{textcomp}%
\usepackage{manyfoot}%
\usepackage{booktabs}%
\usepackage{algorithm}%
\usepackage{listings}%
\usepackage{lipsum}
\usepackage{svg}

\usepackage [english]{babel}
\usepackage [autostyle, english = american]{csquotes}
\MakeOuterQuote{"}
\usepackage{algorithmic}
\usepackage[figurename=Fig.]{caption}
\usepackage{subcaption}

\usepackage{makecell}

\usepackage{tabularx}
\newcolumntype{Y}{>{\centering\arraybackslash}X}

\begin{document}

\title[Article Title]{SFDemorpher: Generalizable Face Demorphing for Operational Morphing Attack Detection}


\author*[1]{\fnm{Raul} \sur{Ismayilov}}\email{raul.ismayilov@utwente.nl}
\author[1]{\fnm{Luuk} \sur{Spreeuwers}}\email{l.j.spreeuwers@utwente.nl}
\affil[1]{\orgdiv{Data Management \& Biometrics}, \orgname{University of Twente}, \orgaddress{\street{Drienerlolaan 5}, \city{Enschede}, \postcode{7512AD}, \country{Netherlands}}}

\abstract{
Face morphing attacks compromise biometric security by creating document images that verify against multiple identities, posing significant risks from document issuance to border control. Differential Morphing Attack Detection (D-MAD) offers an effective countermeasure, particularly when employing face demorphing to disentangle identities blended in the morph. However, existing methods lack operational generalizability due to limited training data and the assumption that all document inputs are morphs. This paper presents SFDemorpher, a framework designed for the operational deployment of face demorphing for D-MAD that performs identity disentanglement within joint StyleGAN latent and high-dimensional feature spaces. We introduce a dual-pass training strategy handling both morphed and bona fide documents, leveraging a hybrid corpus with predominantly synthetic identities to enhance robustness against unseen distributions. Extensive evaluation confirms state-of-the-art generalizability across unseen identities, diverse capture conditions, and 13 morphing techniques, spanning both border verification and the challenging document enrollment stage. Our framework achieves superior D-MAD performance by widening the margin between the score distributions of bona fide and morphed samples while providing high-fidelity visual reconstructions facilitating explainability. 
}


\maketitle


\section{Introduction} \label{sec:introduction}
Face morphing attacks~\cite{magic_passport} compromise the unique identity binding principle in biometric security, creating vulnerabilities exploitable from document issuance to verification. By blending facial features, an accomplice can fraudulently enroll a morphed image to obtain a travel document for a criminal. This criminal can then use the document at an Automated Border Control (ABC) system~\cite{ABC_systems}, where the stored morph is compared to a live capture, potentially allowing unauthorized entry. To counter this, Morphing Attack Detection (MAD)~\cite{MAD} aims to identify such forgeries. MAD includes Single-image MAD (S-MAD), analyzing standalone documents, and Differential MAD (D-MAD), detecting discrepancies by comparing the document against a trusted reference~\cite{MAD_survey}.

Face demorphing~\cite{face_demorphing} is a D-MAD method that reconstructs the second identity contributing to the morph but absent from the trusted reference. Specifically, during the morphed document enrollment stage, reconstruction aims to recover the criminal's identity, while during border control verification, it aims to reconstruct the accomplice's identity. Comparing this reconstruction to the trusted reference via a Face Recognition System (FRS) enables MAD.

However, generalizability to unseen morphing techniques and diverse capture conditions remains a challenge. Moreover, generative models such as GANs~\cite{GANs} and diffusion models~\cite{diffusion_models} now facilitate the creation of high-fidelity morphs~\cite{mipgan, mipgan_improved, ladimo}, necessitating more advanced detection capabilities.

For this reason, recent face demorphing research leverages generative models~\cite{diffusion_demorphing, cnn_demorphing, dcGAN, fdgan, stylegan_demorphing}. However, these methods often suffer from generalizability limitations due to restrictive assumptions about training and testing distributions. Constraints include requiring identical morphing techniques, similar capture conditions, or overlapping identity pools, limiting real-world applicability. The recent diffDeMorph~\cite{diffDemorph} approach addresses this via RGB-domain conditioning and coupled diffusion for improved generalizability across unseen morphing techniques and image styles. Similarly, StyleDemorpher~\cite{styledemorpher} uses pre-trained StyleGAN~\cite{stylegan2} for latent space demorphing, demonstrating generalizability to unseen identities, image conditions, and morphing methods. However, most recent methods neglect D-MAD performance, often assuming all inputs are morphs and treating demorphing solely as an identity decomposition task. Although StyleDemorpher evaluates D-MAD, its performance on bona fide images degrades due to morph-only training.

This paper introduces the SFDemorpher framework, addressing generalizability challenges and focusing on D-MAD performance for operational deployment. Similar to StyleDemorpher, we utilize StyleGAN for image synthesis. However, demorphing occurs in both the expanded latent space $\mathcal{W}^+$~\cite{w_plus_space} and the high-dimensional feature space $\mathcal{F}^k$~\cite{sfe}, improving identity preservation crucial for accurate reconstruction. Moreover, we introduce a dual-pass training procedure using both morphed and bona fide images, improving D-MAD performance and achieving state-of-the-art (SOTA) results. This training leverages a hybrid dataset comprising FLUXSynID~\cite{fluxsynid} and DemorphDB~\cite{styledemorpher}. The corpus consists of 80\% synthetic identities from FLUXSynID and 20\% real identities from DemorphDB. To our knowledge, this is the first face demorphing approach using synthetic data as the majority of its training set.

We extensively evaluate SFDemorpher on image reconstruction and D-MAD performance across three datasets separated from the training corpus. Beyond accomplice restoration, we also focus on the challenging criminal restoration scenario~\cite{iciap2023, acida}, which is largely unexplored. We demonstrate SOTA performance, ensuring generalizability and utility for operational deployment.

The contributions of this paper are as follows: 
\begin{itemize} 
\item \textbf{Joint Style-Feature Face Demorphing:} We propose a novel architecture leveraging StyleGAN's latent and feature spaces. This joint approach significantly improves identity preservation for high-quality demorphing.
\item \textbf{Scenario-Aware Training:} We address the limitation of existing methods assuming all inputs are morphs by introducing a training strategy for both bona fide and morphed documents. This ensures effective identity preservation and disentanglement, improving D-MAD.
\item \textbf{Large-Scale Synthetic Data Utilization:} We overcome data scarcity by training on a hybrid corpus comprising predominantly synthetic identities. To our knowledge, this is the first face demorphing approach relying predominantly on synthetic data, demonstrating effective generalization to real-world scenarios.
\item \textbf{Operational Viability:} We conduct extensive evaluations on unseen identities, capture conditions, and morphing methods. Our results demonstrate SOTA performance, validating SFDemorpher's robustness and potential for real-world deployment.
\end{itemize}

\section{Related Work} \label{sec:related_work}
A face morphing attack combines facial images from two distinct individuals, typically an accomplice who applies for an identity document and a criminal who exploits it, to create a single composite image verifiable against both identities in automated Face Recognition Systems (FRS)~\cite{magic_passport, MAD, MAD_survey}. This threat operates during travel document issuance, where morphed photos bypass human examiner scrutiny~\cite{acida}, and at Automated Border Control (ABC) gates~\cite{ABC_systems} where criminals authenticate using their facial features preserved in the morphs~\cite{face_demorphing}. Early landmark-based morphing methods using Delaunay triangulation~\cite{splicing, 7935087} often required manual artifact removal, limiting scalability. Recent advances include MIPGAN~\cite{mipgan} employing StyleGAN~\cite{stylegan2} with identity-prior-driven loss functions, LADIMO~\cite{ladimo} leveraging latent diffusion for biometric template inversion, and improved automated landmark-based approaches reducing ghosting artifacts~\cite{utw_morphs}.

To counter these threats, Morphing Attack Detection (MAD) techniques are typically categorized into Single-image MAD (S-MAD) and Differential MAD (D-MAD)~\cite{MAD, MAD_survey}. S-MAD analyzes isolated documents for intrinsic artifacts such as frequency domain anomalies~\cite{8846232, Ramachandra2019DetectingFM, TAPIA2025130033, Neubert2019AFM}. Conversely, D-MAD compares suspected documents against trusted live captures to detect identity inconsistencies and artifacts often invisible in single-image analysis~\cite{9093905, siamese, iciap2023}. In~\cite{9912424}, feature-wise supervision with similarity and distance-based losses was employed to localize morphed areas, while in~\cite{11099318} multimodal large language models were utilized for zero-shot D-MAD to improve interpretability. Furthermore, the ACIdA framework~\cite{acida} evaluates D-MAD across criminal and accomplice verification scenarios, noting significant degradation in identity-based methods~\cite{9093905} under the latter scenario. ACIdA addresses this critical limitation by a modular framework combining attempt classification with identity and artifact analysis.

Among D-MAD methods, face demorphing~\cite{face_demorphing} operates differently by not only detecting morphs but also reconstructing an image of the contributor absent from the live capture. Building on prior work in~\cite{magic_passport, Ferrara2016}, this approach manipulates facial landmarks and employs geometric warping to invert the morphing process. Subsequent work shifted to deep learning-based image synthesis~\cite{fdgan, cnn_demorphing, dcGAN, diffDemorph, styledemorpher}. For instance, the work in~\cite{diffusion_demorphing} leverages diffusion autoencoders~\cite{diff_ae} to encode morphs into disentangled latent spaces, where a dual-branch network isolates accomplice features. Similarly, in~\cite{stylegan_demorphing} images are encoded into the StyleGAN~\cite{stylegan2} latent space and a lightweight separation network with cross-attention and residual modules is used to isolate the accomplice identity.

While early approaches~\cite{fdgan, cnn_demorphing} suffered from artifacts and resolution limits, recent works~\cite{diffusion_demorphing, styledemorpher, stylegan_demorphing} leverage pre-trained generative models to mitigate these limitations. However, most methods rely on limited data due to the scarcity of high-quality document and live capture image pairs. Furthermore, training and testing sets often share identical distributions, failing to reflect operational scenarios where unseen during training identities, capture conditions, and morphing methods are prevalent. Recent works~\cite{diffDemorph, styledemorpher} address aspects of this generalizability. Notably, StyleDemorpher~\cite{styledemorpher} introduced the DemorphDB dataset, combining five facial databases (FRGC~\cite{FRGC}, Eurecom-IST~\cite{IST-EURECOM}, Utrecht ECVP~\cite{utrecht_ECVP_dataset}, Chicago Face Database~\cite{chicago_db_1, chicago_db_2, chicago_db_3}, Face Research Lab London Dataset~\cite{frll}). By re-purposing and fine-tuning a StyleGAN~\cite{stylegan2} encoder for demorphing on DemorphDB, the authors demonstrated generalization to unseen identities, morphing methods, and image corruptions.

While StyleDemorpher~\cite{styledemorpher} demonstrates generalizability, it relies exclusively on the $\mathcal{W}^+$ latent space of StyleGAN~\cite{stylegan2}, similar to the work in~\cite{stylegan_demorphing}. Operating in $\mathcal{W}^+$ latent space has been shown to be lossy~\cite{pehlivan2023styleres, 9878687, yao2022featurestyleencoderstylebasedgan}, leading to identity information loss. However, effective demorphing requires balancing inversion quality and editability~\cite{editability_inversion_tradeoff}. In the context of face demorphing, inversion quality refers to faithfully reconstructing identity features from real-world images, critical for preserving identity in documents and trusted reference images. This conflicts with editability, the capacity to semantically modify the representation to recover the concealed identity within a morph~\cite{styledemorpher}. High inversion quality often forces embeddings into low-density latent regions, making them brittle for editing, while prioritizing editability by constraining embeddings to well-behaved regions sacrifices fine-grained details~\cite{editability_inversion_tradeoff, sfe}. StyleFeatureEditor~\cite{sfe} addresses this by encoding into a high-dimensional feature space $\mathcal{F}^k$ for inversion quality, using a separate network for editability.

Another limitation of current face demorphing methods is the neglect of bona fide documents, given that the majority of document images in operational settings are expected to be bona fide. While the original demorphing method~\cite{face_demorphing} targeted both bona fide and morphed images, subsequent research~\cite{fdgan, diffDemorph, diffusion_demorphing, dcGAN, stylegan_demorphing} assumes all input images are morphs, consequently failing to report D-MAD performance. Although D-MAD performance is evaluated in~\cite{styledemorpher}, the results indicate that training exclusively on morphs biases the model, degrading bona fide identity preservation. Finally, the challenging accomplice verification scenario~\cite{acida}, which corresponds to criminal identity restoration for face demorphing algorithms, has not been explicitly evaluated in current literature, with the majority of demorphing methods focused solely on accomplice restoration.

\section{Methodology}\label{sec:method}
This section details the proposed SFDemorpher framework, designed for high-fidelity face demorphing and robust Differential Morphing Attack Detection (D-MAD). We first formalize operational deployment limitations and define notation for distinct restoration scenarios (Sec.~\ref{sec:problem_formulation}). Next, we present SFDemorpher, utilizing a dual-pass training strategy and joint StyleGAN~\cite{stylegan2} latent and feature spaces (Sec.~\ref{sec:framework}). We then detail the loss functions enabling optimization for both bona fide and morphed images (Sec.~\ref{sec:losses}). Finally, we discuss integrating the framework into a practical D-MAD pipeline (Sec.~\ref{sec:dmad_integration}).

\subsection{Problem Formulation} \label{sec:problem_formulation}
Operational deployment of face demorphing, such as in Automated Border Control (ABC) systems~\cite{ABC_systems}, introduces challenges inadequately addressed in current literature: the generative model inversion quality-editability trade-off and insufficient generalization for operational D-MAD.

\subsubsection{The Inversion Quality-Editability Trade-off}
Prior works employing StyleGAN~\cite{styledemorpher, stylegan_demorphing} use the expanded $\mathcal{W}^+$ latent space, leading to information loss~\cite{pehlivan2023styleres, 9878687, yao2022featurestyleencoderstylebasedgan}. SFDemorpher adapts StyleFeatureEditor~\cite{sfe} to perform inversion into both the $\mathcal{W}^+$ latent space and the $\mathcal{F}^k$ feature space (outputs of the $k$-th convolutional layer of StyleGAN), supporting near-perfect reconstruction. While editing in feature space is difficult, StyleFeatureEditor demonstrates that a dedicated network can be trained to edit the feature maps. SFDemorpher adapts the StyleFeatureEditor inverter network to encode input images into both spaces, ensuring identity preservation, and employs specialized modules for face demorphing within these representations.

\begin{figure*}[htbp]
     \centering
     \begin{subfigure}[htbp]{0.32\linewidth}
         \centering
         \includegraphics[width=0.8\linewidth]{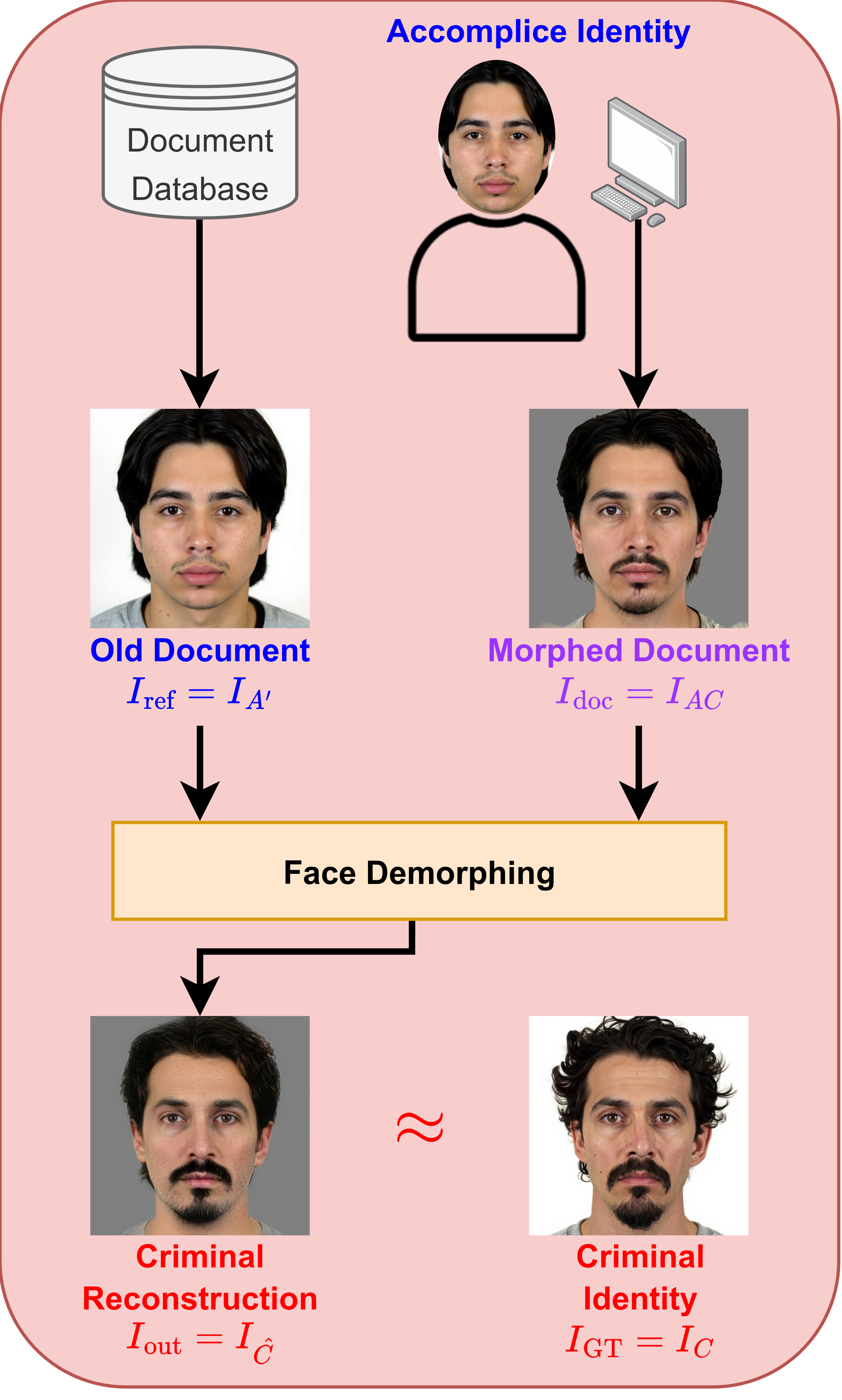}
         \caption{Criminal identity restoration}
         \label{fig:criminal_restoration}
     \end{subfigure}
     \hfill
     \begin{subfigure}[htbp]{0.32\linewidth}
         \centering
         \includegraphics[width=0.8\linewidth]{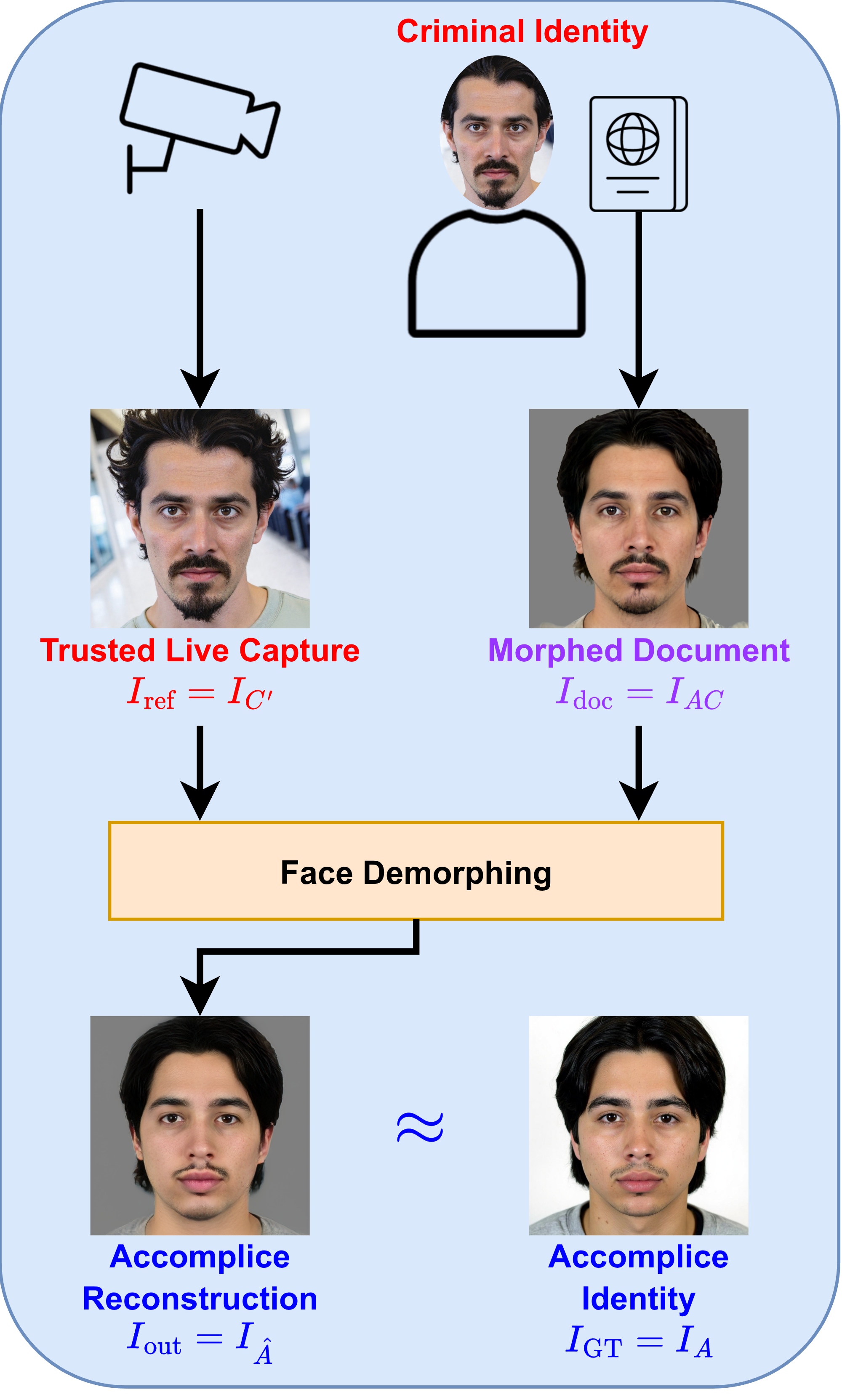}
         \caption{Accomplice identity restoration}
         \label{fig:accomplice_restoration}
     \end{subfigure}
     \hfill
     \begin{subfigure}[htbp]{0.32\linewidth}
         \centering
         \includegraphics[width=0.8\linewidth]{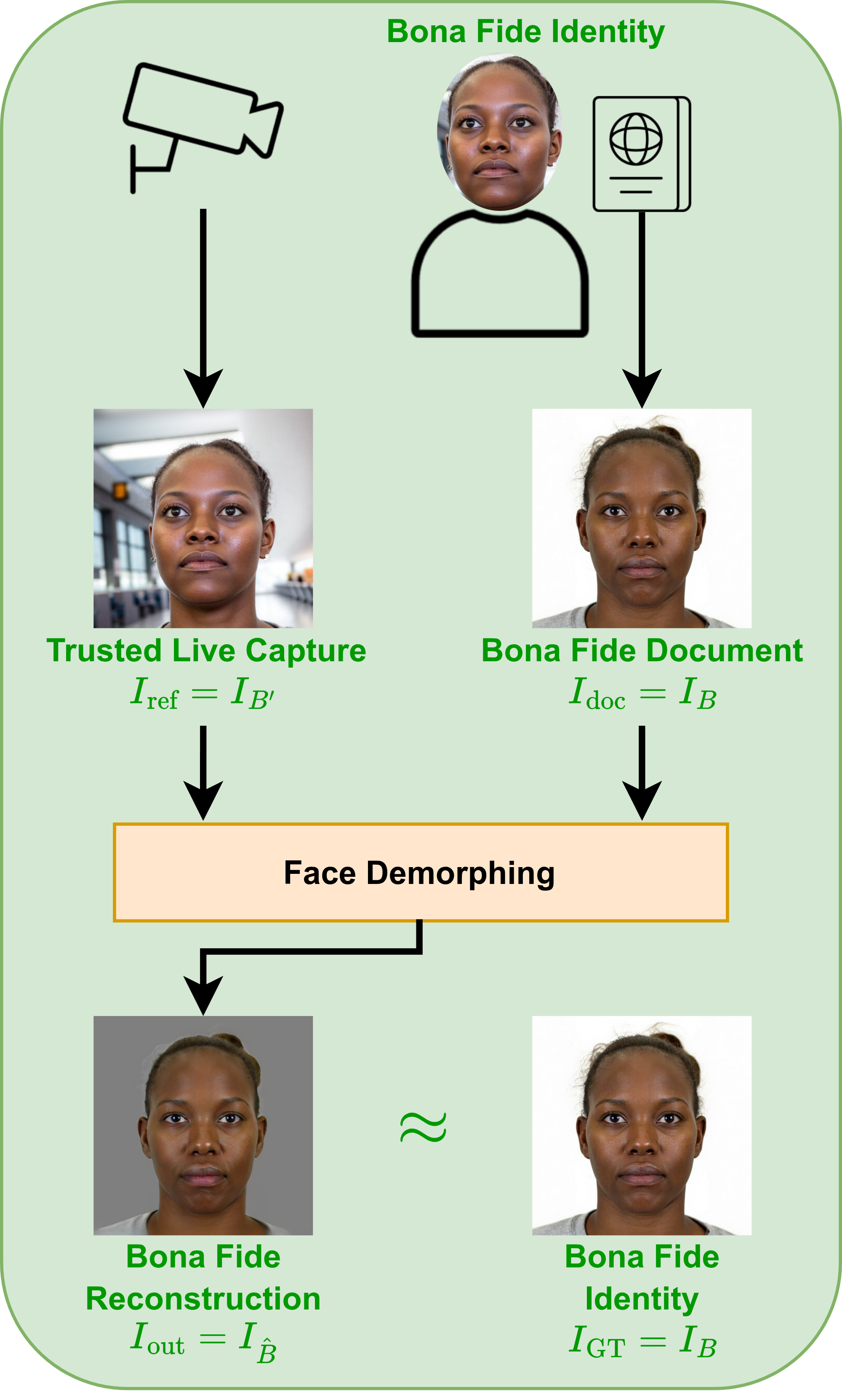}
         \caption{Bona fide identity restoration}
         \label{fig:bonafide_restoration}
     \end{subfigure}
    \caption{Visualization of the three primary operational face demorphing scenarios. In each case, face demorphing algorithm processes a suspected document image $I_{\text{doc}}$ and a trusted reference $I_{\text{ref}}$ to estimate a ground truth $I_{\text{GT}}$. The specific restoration goal $I_{\text{out}}$ is determined by the scenario: \textbf{(a)} fraudulent enrollment with a morphed document, \textbf{(b)} border crossing with a morphed document, and \textbf{(c)} standard travel with a bona fide document.}
     \label{fig:restoration_cases}
\end{figure*}

\subsubsection{Operational Generalization}
A critical barrier to real-world deployment of face demorphing is limited generalizability to unseen data distributions and the assumption that all input documents are morphs. Robustness against unseen identities, capture conditions, and morphing methods requires high data variability, which is difficult to obtain. While morphed data can be expanded combinatorially, scaling bona fide datasets (paired document and live images) is challenging. SFDemorpher addresses this via a specialized training strategy. We leverage the FLUXSynID~\cite{fluxsynid} synthetic dataset, comprising identities with paired document and live capture images, as the majority of our training corpus. Its size and diversity enables learning representations that transfer to real-world scenarios.

\subsubsection{Formal Definitions and Restoration Scenarios}
We define $I_A$ and $I_C$ as original accomplice and criminal document images, $I_B$ as a bona fide document, and $I_{AC}$ as a morph generated using $I_A$ and $I_C$. Prime notation (e.g., $I_{A'}$) denotes a trusted reference image (e.g., live capture), and hat notation (e.g., $I_{\hat{A}}$) denotes the reconstructed output of the face demorphing algorithm.

In a generalized context, face demorphing processes a suspected document image $I_{\text{doc}}$ and a trusted reference $I_{\text{ref}}$ to produce a restored image $I_{\text{out}}$ estimating a ground truth identity $I_{\text{GT}}$. We identify three restoration scenarios (see Fig.~\ref{fig:restoration_cases}):

\begin{itemize}
    \item \textbf{Criminal Identity Restoration (Fig.~\ref{fig:criminal_restoration}):} During document enrollment, an accomplice submits a morph ($I_{\text{doc}} = I_{AC}$). The face demorphing algorithm compares it against a trusted accomplice reference ($I_{\text{ref}} = I_{A'}$) to reconstruct the concealed criminal identity ($I_{\text{out}} = I_{\hat{C}}$). The ground truth is the original criminal image used to generate the morph ($I_{\text{GT}} = I_C$). Success prevents fraudulent document issuance.
    \item \textbf{Accomplice Identity Restoration (Fig.~\ref{fig:accomplice_restoration}):}
    Here, a criminal uses an issued morphed document ($I_{\text{doc}} = I_{AC}$) at an ABC gate. The system captures a live image of the criminal ($I_{\text{ref}} = I_{C'}$) and aims to reconstruct the missing accomplice identity ($I_{\text{out}} = I_{\hat{A}}$). The ground truth is the original accomplice image ($I_{\text{GT}} = I_A$). Success prevents illegal entry.
    \item \textbf{Bona Fide Identity Restoration (Fig.~\ref{fig:bonafide_restoration}):} This represents standard travel with a bona fide document ($I_{\text{doc}} = I_{B}$) and a matching trusted reference ($I_{\text{ref}} = I_{B'}$). The goal is faithful reconstruction ($I_{\text{out}} \approx I_{\text{doc}}$), with the ground truth being the document itself ($I_{\text{GT}} = I_{B}$). This ensures integrity and prevents artifacts when no morphing is present.
\end{itemize}

\begin{figure*}[htbp]
    \centering
    \includegraphics[width=0.99\linewidth]{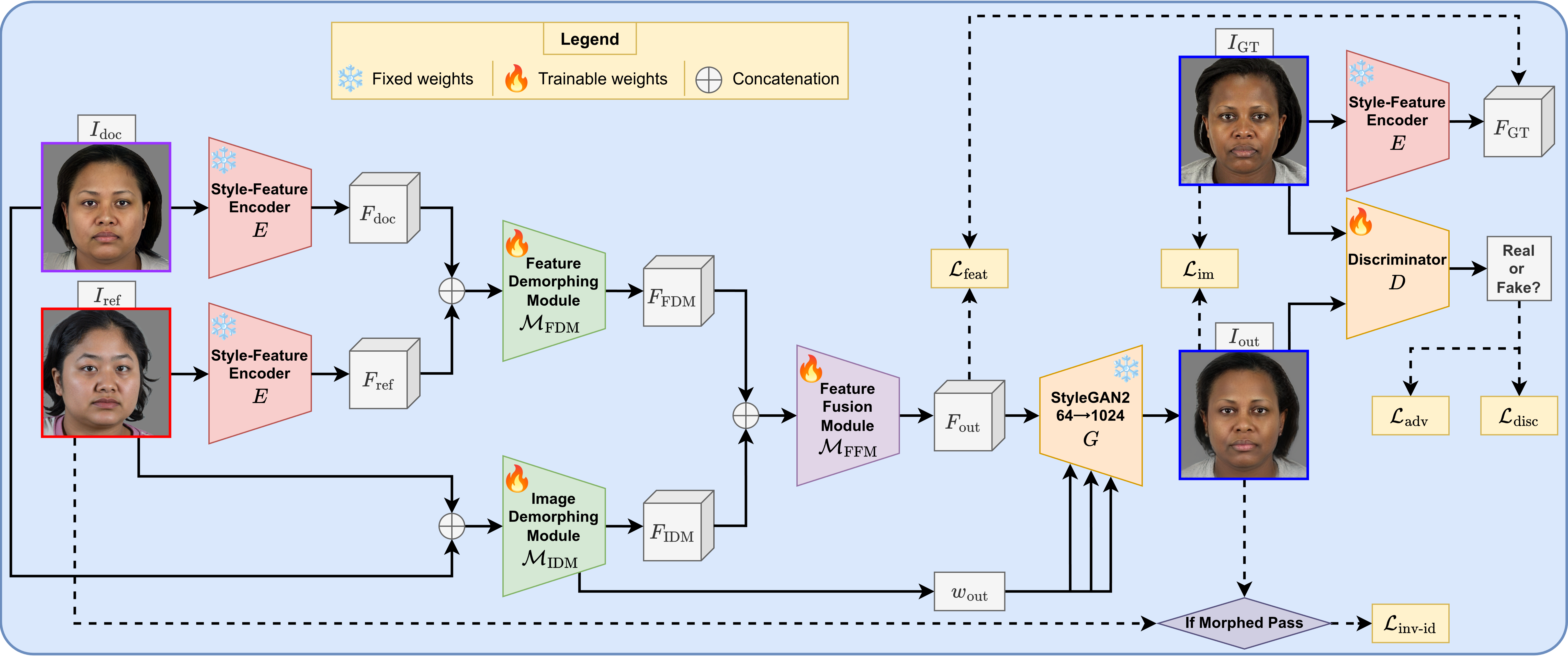}
    \caption{The training pipeline of the SFDemorpher framework utilizing a dual-pass strategy. The \textbf{bona fide pass} processes the input pair $(I_{\text{doc}}, I_{\text{ref}}) = (I_B, I_{B'})$ to reconstruct the ground truth $I_{\text{GT}} = I_B$, yielding output $I_{\text{out}} = I_{\hat{B}}$. The \textbf{morphed pass} trains on the accomplice restoration scenario using inputs $(I_{\text{doc}}, I_{\text{ref}}) = (I_{AC}, I_{C'})$ to disentangle identities, targeting the original accomplice $I_{\text{GT}} = I_A$ to reconstruct $I_{\text{out}} = I_{\hat{A}}$. During inference, the framework operates identically without the ground truth $I_{\text{GT}}$.}
    \label{fig:pipeline}
\end{figure*}

Criminal identity restoration is inherently more challenging than accomplice restoration~\cite{acida}. To maximize verification success during the document application, high-quality splicing-based~\cite{splicing} morphs typically embed the morphed inner face into the outer facial region of the accomplice. Consequently, the resulting morph entirely lacks the criminal's outer facial data, making full criminal identity reconstruction a severely ill-posed task.

\subsection{SFDemorpher Framework} \label{sec:framework}
The proposed SFDemorpher framework utilizes a joint StyleGAN~\cite{stylegan2} style-feature representation~\cite{sfe} for high-fidelity face demorphing of bona fide and morphed documents. The complete training pipeline is visualized in Fig.~\ref{fig:pipeline}.

SFDemorpher accepts both a suspected travel document image $I_{\text{doc}}$ and a trusted reference $I_{\text{ref}}$. All input images are preprocessed using the pre-trained BiRefNet~\cite{birefnet} network for background removal, aligned using the FFHQ~\cite{stylegan1} protocol, and resized to $256 \times 256$.

The architecture processes inputs via two parallel pathways. In the first pathway, a frozen Style-Feature Encoder $E$~\cite{sfe} maps images to both the expanded latent space $\mathcal{W}^+$ and the high-dimensional feature space $\mathcal{F}^k$:
\begin{equation}
\begin{aligned}
    \left( w_{\text{doc}}, F_{\text{doc}}\right) &= E\left( I_{\text{doc}}\right), \\
    \left( w_{\text{ref}}, F_{\text{ref}}\right) &= E\left( I_{\text{ref}}\right),
\end{aligned}
\label{eq:fse_encoding}
\end{equation}
\noindent where $w \in \mathcal{W^+}$ is $\mathbb{R}^{18\times512}$, and $F \in \mathcal{F}^9$ is $\mathbb{R}^{512\times64\times64}$, corresponding to the feature maps of the 9-th convolutional layer of StyleGAN. In this pathway, the latent codes $\left( w_{\text{doc}}, w_{\text{ref}} \right)$ are discarded to solely leverage the rich spatial information embedded in the feature maps. The resulting feature maps $\left( F_{\text{doc}}, F_{\text{ref}} \right)$ are concatenated along the channel dimension and processed by the Feature Demorphing Module, $\mathcal{M}_{\text{FDM}}$:
\begin{equation}
F_{\text{FDM}} = \mathcal{M}_{\text{FDM}}\left( F_{\text{doc}} \mathbin\Vert F_{\text{ref}}\right),
\label{eq:feature_demorphing_module}
\end{equation}
\noindent where $\mathbin\Vert$ denotes channel-wise concatenation. The output is a feature map $F_{\text{FDM}}\in\mathbb{R}^{512\times64\times64}$.

The second pathway concatenates $I_{\text{doc}}$ and $I_{\text{ref}}$ channel-wise for the Image Demorphing Module, $\mathcal{M}_{\text{IDM}}$. Unlike $\mathcal{M}_{\text{FDM}}$, this module operates directly on the pixel domain:
\begin{equation}
\left(w_{\text{out}}, F_{\text{IDM}} \right) = \mathcal{M}_{\text{IDM}}\left( I_{\text{doc}} \mathbin\Vert I_{\text{ref}} \right),
\label{eq:image_demorphing_module}
\end{equation}
\noindent where $w_{\text{out}}\in\mathbb{R}^{18\times512}$ represents the demorphed latent code and $F_{\text{IDM}}\in\mathbb{R}^{512\times64\times64}$ represents the corresponding demorphed feature maps.

Next, the Feature Fusion Module, $\mathcal{M}_{\text{FFM}}$, learns to selectively aggregate optimal spatial features from both parallel pathways:
\begin{equation}
F_{\text{out}} = \mathcal{M}_{\text{FFM}}\left( F_{\text{FDM}} \mathbin\Vert F_{\text{IDM}}\right),
\label{eq:feature_fusion_module}
\end{equation}
\noindent yielding the final fused demorphed feature map $F_{\text{out}}\in\mathbb{R}^{512\times64\times64}$.

Finally, the frozen StyleGAN generator, $G$, synthesizes the image. As synthesis begins at the 9-th layer ($k=9$) via $F_{\text{out}}$ injection, only subsequent latent codes are required:
\begin{equation}
I_{\text{out}} = G\left( F_{\text{out}}, w_{\text{out}}^{k+1:N} \right),
\label{eq:image_synthesis}
\end{equation}
\noindent where $N$ is the total number of StyleGAN layers and $w_{\text{out}}^{k+1:N}$ denotes the latent codes $w_{\text{out}}^{k+1}, \dots, w_{\text{out}}^{N}$. The resulting output $I_{\text{out}}$ aims to match the ground truth identity image $I_{\text{GT}}$, guided by the loss functions detailed in Sec.~\ref{sec:losses}.

Distinct from other methods~\cite{styledemorpher, cnn_demorphing, diffusion_demorphing, dcGAN, fdgan, diffDemorph, stylegan_demorphing}, we employ a dual-pass training strategy for operational generalizability. The bona fide pass learns faithful reconstruction $(I_{\text{doc}}, I_{\text{ref}}, I_{\text{GT}}, I_{\text{out}}) = (I_{B}, I_{B'}, I_{B}, I_{\hat{B}})$, corresponding to the bona fide restoration scenario (Fig.~\ref{fig:bonafide_restoration}). The morphed pass learns identity disentanglement by setting $(I_{\text{doc}}, I_{\text{ref}}, I_{\text{GT}}, I_{\text{out}}) = (I_{AC}, I_{C'}, I_{A}, I_{\hat{A}})$, corresponding to the accomplice restoration scenario (Fig.~\ref{fig:accomplice_restoration}). These passes alternate during training to ensure balanced optimization.

We exclude criminal restoration (Fig.~\ref{fig:criminal_restoration}) from training. As noted in Sec.~\ref{sec:problem_formulation}, splicing-based morphs~\cite{splicing} lack the criminal's outer facial data, rendering reconstruction ill-posed and training unstable. However, while excluded from training, we still evaluate our method on this scenario.

During inference, SFDemorpher employs the same forward pass to predict $I_{\text{out}}$ from $(I_{\text{doc}}, I_{\text{ref}})$. Because this phase reflects real-world operational conditions, the ground truth target $I_{\text{GT}}$ is naturally unavailable. Thus, the loss functions in Fig.~\ref{fig:pipeline} are computed exclusively during training, as formulated in Sec.~\ref{sec:losses}.

\subsection{Loss Function Formulation} \label{sec:losses}
SFDemorpher is optimized via a weighted loss combination with pass-specific loss coefficients (see Tab.~\ref{tab:loss_coeffs}) to balance bona fide preservation and morphed disentanglement. We formulate loss terms below for a single input pair; during training, these are averaged over the mini-batch.

\subsubsection{Image Reconstruction Objectives}
To ensure high-fidelity restoration of the ground truth $I_{\text{GT}}$, we aggregate four distinct objectives into a composite image reconstruction loss, $\mathcal{L}_{\text{im}}$.

\textbf{Pixel and Structural Consistency.} To capture low-frequency details and enforce structural fidelity, we employ a combination of the $L_2$ distance and the Multi-Scale Structural Similarity (MS-SSIM) loss~\cite{ms-ssim}:
\begin{equation}
    \mathcal{L}_{\text{L}_2} = \left \| I_{\text{out}} - I_{\text{GT}} \right \|_2^2,
\label{eq:l2_loss}
\end{equation}
\begin{equation}
    \mathcal{L}_{\text{ms-ssim}} = 1 - \text{MS-SSIM} \left( I_{\text{out}}, I_{\text{GT}} \right).
\label{eq:mssim_loss}
\end{equation}

\textbf{Perceptual Loss.} To enforce consistency in human-perceptible details and texture, we minimize the perceptual LPIPS~\cite{lpips} loss using a pre-trained VGG network $V$~\cite{vgg}:
\begin{equation}
    \mathcal{L}_{\text{lpips}} = \left \|V \left( I_{\text{out}} \right) - V \left( I_{\text{GT}} \right) \right \|_2^2.
\label{eq:lpips_loss}
\end{equation}

\textbf{Identity Loss.} 
To preserve biometric identity information crucial for D-MAD, we leverage a pre-trained AdaFace~\cite{adaface} Face Recognition System (FRS). We define identity loss using the similarity metric $S(\cdot, \cdot)$, measuring the cosine similarity between AdaFace embeddings of $I_{\text{out}}$ and $I_{\text{GT}}$:
\begin{equation}
    \mathcal{L}_{\text{id}} = 1 - S \left(I_{\text{out}}, I_{\text{GT}} \right).
\label{eq:id_loss}
\end{equation}

These four loss terms are combined to form the composite image-level loss:
\begin{equation}
\begin{split}
    \mathcal{L}_{\text{im}} &= \lambda_{\text{L2}}\mathcal{L}_{\text{L2}} + \lambda_{\text{lpips}}\mathcal{L}_{\text{lpips}} \\
    &+ \lambda_{\text{ms-ssim}}\mathcal{L}_{\text{ms-ssim}} + \lambda_{\text{id}}\mathcal{L}_{\text{id}}.
\end{split}
\label{eq:im_loss}
\end{equation}

\subsubsection{Identity Disentanglement and Feature Regularization}
Beyond standard image reconstruction, we employ specific losses to enforce identity separation and regularize the feature space.

\textbf{Inverse Identity Loss.} For the morphed pass specifically, we introduce an Inverse Identity Loss to suppress the criminal identity, preventing feature leakage into the reconstructed accomplice image. It penalizes similarity between $I_{\text{out}}$ and the criminal reference $I_{\text{ref}}$ if it exceeds a margin $m$:
\begin{equation}
    \mathcal{L}_{\text{inv-id}} = \max \left( 0, S \left( I_{\text{out}}, I_{\text{ref}}\right) - m \right),
\label{eq:inv_id_loss}
\end{equation}

\noindent where we set $m=-0.5$ for sufficient separation.

\textbf{Feature Loss.} To ensure consistency in the StyleGAN feature space, we minimize the difference between the intermediate features of the demorphed output $F_{\text{out}}$ and the target features $F_{\text{GT}}$ extracted by the frozen encoder $E$ (see Fig.~\ref{fig:pipeline}). This loss also acts as a regularizer, preventing the demorphing modules from generating unbounded feature representations by anchoring them to the well-behaved feature statistics of the ground truth:
\begin{equation}
    \mathcal{L}_{\text{feat}} = \left \| F_{\text{out}} - F_{\text{GT}} \right \|_2^2.
\label{eq:feat_loss}
\end{equation}

\subsubsection{Adversarial Training}
To ensure the generated images are photo-realistic, we employ an adversarial training setup. It is important to note that the StyleGAN generator $G$ is frozen, and the adversarial loss is used to update the weights of the trainable demorphing modules ($\mathcal{M}_{\text{FDM}}, \mathcal{M}_{\text{IDM}}, \mathcal{M}_{\text{FFM}}$).

\textbf{Generator Objective.} The demorphing modules minimize the non-saturating adversarial loss~\cite{GANs} against a trainable discriminator $D$~\cite{stylegan2}:
\begin{equation}
    \mathcal{L}_{\text{adv}} = -\log \left( D \left( I_{\text{out}} \right)\right).
\label{eq:adv_loss}
\end{equation}

\textbf{Discriminator Objective.} The discriminator $D$ is trained separately to distinguish between real ground truth images and the demorphed reconstructions. Its training is regularized by the $R_1$ gradient penalty~\cite{r1_regularization}:
\begin{equation}
\begin{split}
    \mathcal{L}_{\text{disc}} &= -\log \left( D\left( I_{\text{GT}} \right)\right) - \log \left( 1 - D\left( I_{\text{out}}\right)\right) \\
    &+ \frac{\gamma}{2} \left \| \nabla_{I_{\text{GT}}} D \left( I_{\text{GT}} \right)\right \|_2^2,
\end{split}
\label{eq:disc_loss}
\end{equation}
where $\gamma$ scales the gradient penalty strength and is set to $10$, following~\cite{sfe}.

\subsubsection{Total Training Objective}
The final objective function optimizes the trainable demorphing modules by combining the image reconstruction loss with the disentanglement, regularization, and adversarial constraints:
\begin{equation}
\begin{split}
    \mathcal{L}_{\text{total}} &= \mathcal{L}_{\text{im}} + \lambda_{\text{inv-id}}\mathcal{L}_{\text{inv-id}} \\
    &+ \lambda_{\text{feat}}\mathcal{L}_{\text{feat}} + \lambda_{\text{adv}}\mathcal{L}_{\text{adv}}.
\end{split}
\label{eq:total_loss}
\end{equation}

Table~\ref{tab:loss_coeffs} details the selected loss coefficients. Starting from the recommendations in~\cite{sfe, styledemorpher}, the final coefficients were established through an extensive empirical search across varying magnitudes. The morphed pass utilizes significantly higher coefficient values to guide the optimization through the complex non-linear identity disentanglement process, whereas the bona fide pass uses lower weights to prioritize subtle preservation.

\begin{table}[htbp]
\centering
\footnotesize
\caption{Loss coefficients ($\lambda$) for the dual-pass training strategy. Note that $\mathcal{L}_{\text{inv-id}}$ is only active during the morphed pass.}
\label{tab:loss_coeffs}
\begin{tabular}{lcc}
\hline
\textbf{Coefficient} & \textbf{Bona Fide Pass} & \textbf{Morphed Pass} \\ \hline
$\lambda_{\text{id}}$       & 0.1  & 1.0  \\
$\lambda_{\text{L2}}$       & 0.1  & 1.0  \\
$\lambda_{\text{lpips}}$    & 0.08 & 0.8  \\
$\lambda_{\text{ms-ssim}}$  & 0.04 & 0.4  \\
$\lambda_{\text{feat}}$     & 0.01 & 0.1  \\
$\lambda_{\text{inv-id}}$   & 0.0  & 0.6  \\
$\lambda_{\text{adv}}$      & 0.01 & 0.01 \\ \hline
\end{tabular}
\end{table}

\subsection{D-MAD Integration} \label{sec:dmad_integration}
\begin{figure}[htbp]
    \centering
    \includegraphics[width=\linewidth]{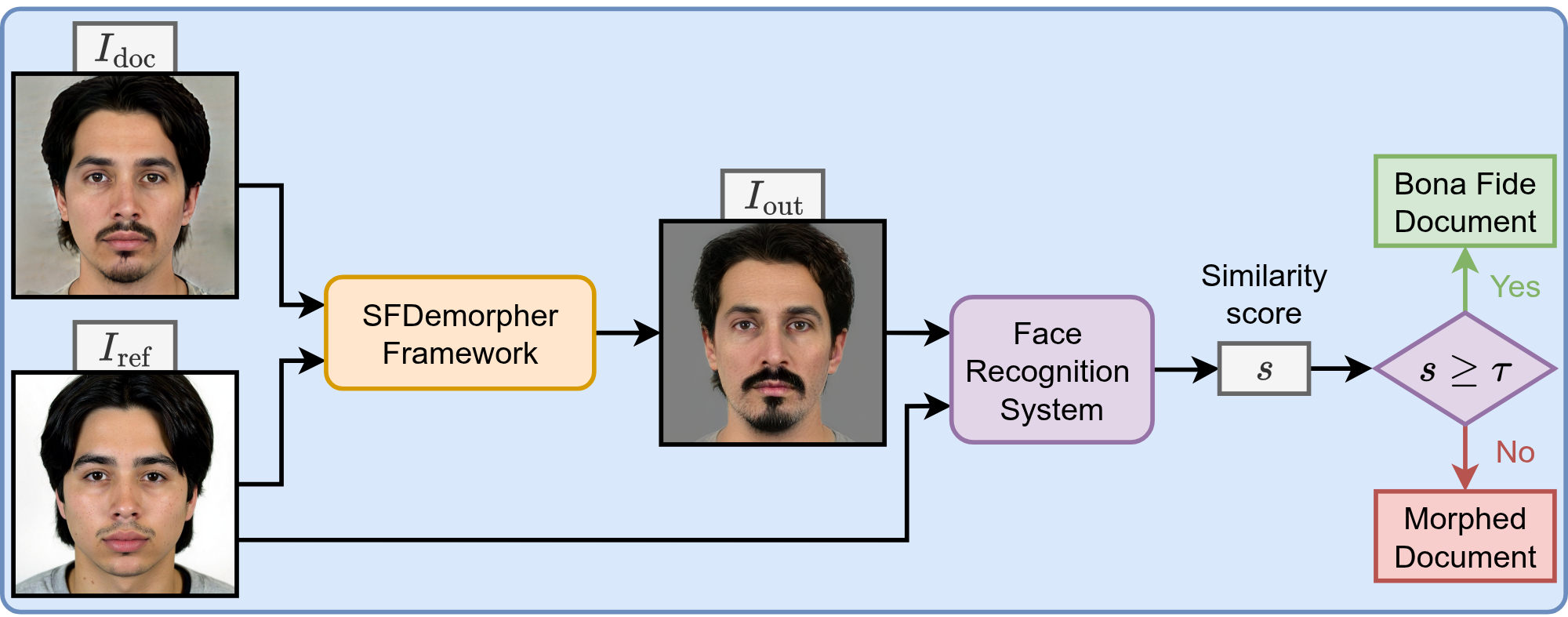}
    \caption{Overview of the proposed D-MAD pipeline. The SFDemorpher framework reconstructs an output image $I_{\text{out}}$ from the document $I_{\text{doc}}$ and trusted reference $I_{\text{ref}}$. Next, an FRS computes the similarity score $s$ between the reconstruction and the reference to classify the document as either bona fide or morphed.} 
    \label{fig:dmad_setup}
\end{figure}

As a generative framework, SFDemorpher reconstructs facial images rather than directly performing D-MAD. Utilization for D-MAD requires integration with an FRS for decision-making~\cite{face_demorphing}. Our D-MAD pipeline is illustrated in Fig.~\ref{fig:dmad_setup}.

The suspected document $I_{\text{doc}}$ and trusted reference $I_{\text{ref}}$ fed into the trained SFDemorpher yield reconstruction $I_{\text{out}}$. Next, an FRS (in this work, we employ AdaFace~\cite{adaface}) extracts identity embeddings for $I_{\text{out}}$ and $I_{\text{ref}}$. The decision score $s$ is the cosine similarity between these embeddings:

\begin{equation}
    s = S \left( I_{\text{ref}}, I_{\text{out}} \right).
\label{eq:similarity_score}
\end{equation}

Classification relies on the premise that successful demorphing of bona fide documents yields reconstructions highly similar to the reference (high score). Conversely, for morphed documents, the objective is reconstructing the concealed identity distinct from the reference (low score). Thus, classification $C$ compares $s$ against threshold $\tau$:

\begin{equation}
    C(I_{\text{doc}}) = 
    \begin{cases} 
    \text{Bona Fide}, & \text{if } s \geq \tau \\
    \text{Morphed}, & \text{if } s < \tau
    \end{cases}.
\label{eq:decision_logic}
\end{equation}

This strategy offers an advantage over traditional D-MAD methods that often, explicitly or implicitly, detect low-level artifacts such as ghosting or GAN spectral fingerprints~\cite{acida, siamese, iciap2023, spectral_dmad}. Performance of such methods can degrade against unseen morphing methods where artifacts are absent or vary. SFDemorpher reframes the task from artifact detection to biometric consistency via identity disentanglement, becoming agnostic to the morphing process. This ensures robustness against high-fidelity, unseen attacks, even if obvious visual anomalies might be ignored, as the core principle of identity disentanglement remains.

Further implementation details, including module architectures and training hyperparameters, are detailed in Appendix~\ref{appendix:implementation_details}.

\section{Experiments} \label{sec:experiments}
This section evaluates SFDemorpher against existing face demorphing and D-MAD methods. Section~\ref{sec:dataset_details} details the datasets. Section~\ref{sec:evaluation_metrics} defines metrics, followed by the experimental setup in Sec.~\ref{sec:experimental_protocol}. Finally, results are presented in Sec.~\ref{sec:evaluation}.

\subsection{Datasets} \label{sec:dataset_details}
\subsubsection{Training Datasets}
SFDemorpher is trained on a hybrid dataset comprising 80\% synthetic and 20\% real data. The model learns from diverse real and synthetic identities across both landmark-based and deep learning-based morphs, ensuring generalizability.

\textbf{Real Data (DemorphDB).} We utilize the DemorphDB dataset~\cite{styledemorpher}, which contains 1,653 real identities. It includes three morphing methods, of which we employ two: UTW-NS~\cite{styledemorpher} (landmark-based) and UTW-StyleGAN~\cite{styledemorpher} (deep learning-based), each with 36,983 morphs. Morphs were created by matching subjects with multiple look-alikes across various source images, providing up to 50 morphs per demorphing reconstruction target for each morphing method. We exclude the third method, UTW~\cite{utw_morphs}, as it explicitly swaps facial components (e.g., eyes, nose), causing irreversible information loss that creates an ill-posed reconstruction target and destabilizes training.

\textbf{Synthetic Data (FLUXSynID).} Most of our training data stems from FLUXSynID~\cite{fluxsynid}, providing 14,889 synthetic identities with paired document and live capture images. Identities with glasses or headwear are excluded to prevent generation of morphing artifacts. For remaining subjects, we generate six morphs using two protocols:
\begin{enumerate} 
\item \textbf{Random Pairing:} Three morphs pair the subject with random identities of the same gender. These pairs exhibit large facial differences, targeting high-level identity disentanglement. 
\item \textbf{Look-alike Pairing:} Three morphs pair the subject with distinct identities exhibiting high facial similarity. Specifically, we select from the pool of five non-mated subjects closest to the AdaFace $0.1\%$ False Match Rate (FMR) decision threshold, providing samples that train the model to handle subtle demorphing tasks.
\end{enumerate} 
This procedure is applied using both UTW-NS and UTW-StyleGAN, resulting in 12 morphs per identity. Finally, we remove low-quality morphs which do not successfully verify against both contributing subjects, resulting in 75,810 UTW-NS and 78,569 UTW-StyleGAN morphs.

\subsubsection{Evaluation Datasets}
To assess operational generalizability, we evaluate on three real-world datasets strictly separated from training, ensuring zero overlap in identities and capture conditions. Evaluation encompasses 13 morphing algorithms, only one of which was used during training (UTW-StyleGAN).

\textbf{FRLL-Morphs-UTW}~\cite{frll_morphs_1, frll_morphs_2, styledemorpher}\textbf{.} Derived from the Face Research Lab London (FRLL) dataset~\cite{frll}, this dataset contains 102 identities. We select neutral expression images as bona fide document samples and smiling expression images as trusted references. We utilize the provided OpenCV~\cite{opencv_morphs}, FaceMorpher~\cite{facemorpher}, WebMorph~\cite{webmorph}, and AMSL~\cite{amsl} morphs, but replace the low-quality StyleGAN~\cite{stylegan2} morphs with high-quality UTW~\cite{utw_morphs} and UTW-StyleGAN~\cite{styledemorpher} morphs. Excluding UTW-StyleGAN, all morphing methods in this data are unseen during training, and the evaluation is limited to accomplice restoration.

\textbf{HNU-FM}~\cite{hnu_fm}\textbf{.} This dataset comprises multiple images of 63 identities with expression and occlusion variations. For bona fide document samples, we select only high-quality images with neutral expressions, open eyes, and no glasses. For trusted references, we sample from the entire image set, including images with varying expressions and occlusions. The dataset employs a single landmark-based morphing method and focuses exclusively on the accomplice restoration scenario.

\textbf{FEI Morph V2}~\cite{iciap2023, feimorph}\textbf{.} Utilizing the 200 identities from the FEI Face Database~\cite{fei}, we employ the single neutral image per identity as the bona fide document sample, while selecting two specific images with varying expression and illumination as trusted references. The dataset includes seven morphing methods (C01~\cite{facemorpher}, C02~\cite{facefusion}, C03~\cite{MAD}, C05~\cite{face_demorphing}, C08, C15~\cite{ingroupe_website, surys_website}, C16~\cite{visualizing_traces}) with 2,000 morphs each. We exclude C01 morphs (FaceMorpher~\cite{MAD}) as they lack splicing~\cite{splicing} and duplicate a morphing method in FRLL-Morphs-UTW. The remaining six splicing-based methods ensure consistency and enable evaluation of both criminal and accomplice restoration scenarios.

We evaluate the quality of all morphing methods in the evaluation datasets in Appendix~\ref{appendix:map}.

\subsection{Evaluation Metrics} \label{sec:evaluation_metrics}



\subsubsection{Identity Deviation}
To evaluate identity consistency across restoration scenarios, we introduce two metrics: Deviation of Target Identity (DTI) and Deviation of Non-Target Identity (DNTI). These generalize DAI, DCI, and DLI~\cite{diffusion_demorphing, styledemorpher} by quantifying the identity features retained or removed during demorphing. 

We define $X \in \{A, C, B\}$ as the target reconstruction identity. In morphing scenarios, $\bar{X}$ denotes the complementary non-target identity in the trusted reference. Specifically, when the target identity is the accomplice ($X=A$), the non-target is the criminal ($\bar{X}=C$), and vice versa.

\textbf{Deviation of Target Identity (DTI).}
This metric quantifies how strongly the intended target identity $X$ is preserved in the reconstruction. Here, $I_{\hat{X}}$ represents the demorphed output image generated to recover $X$, and $N$ is the total number of document-reference pairs in the evaluation set of the corresponding restoration scenario: 
\begin{equation}
    \text{DTI}(X) = \frac{1}{N} \sum_{i=1}^{N} \left( S\left(I_{\hat{X}}^{\left(i\right)}, I_{X}^{\left(i\right)}\right) - \tau \right).
    \label{eq:DTI}
\end{equation}

\noindent A high, positive DTI$(X)$ indicates the target identity is present in the output. We report DTI$(C)$, DTI$(A)$, and DTI$(B)$ for criminal, accomplice, and bona fide restoration, respectively.

\textbf{Deviation of Non-Target Identity (DNTI).}
DNTI evaluates disentanglement by measuring residual traces of the non-target identity $\bar{X}$ within the reconstruction $I_{\hat{X}}$. Here, $I_{\bar{X}'}$ is the trusted reference of the non-target identity:
\begin{equation}
    \text{DNTI}(X) = \frac{1}{N} \sum_{i=1}^{N} \left( S\left(I_{\hat{X}}^{\left(i\right)}, I_{\bar{X}'}^{\left(i\right)}\right) - \tau \right).
    \label{eq:DNTI}
\end{equation}

\noindent A low, negative DNTI$(X)$ implies effective removal of $\bar{X}$. We report DNTI$(A)$ and DNTI$(C)$ for accomplice and criminal restoration. This metric is not applicable to bona fide documents.

\subsubsection{D-MAD Metrics}
We assess D-MAD performance using standardized metrics defined in~\cite{iso20059}. Let $N_{M}$ and $N_{B}$ be the number of morphed and bona fide document-reference image pairs, respectively. The decision is based on the similarity score $s = S(I_{\text{out}}, I_{\text{ref}})$ and an FRS decision threshold $\tau$ (see Fig.~\ref{fig:dmad_setup}).

\textbf{Morphing Attack Classification Error Rate (MACER).}
MACER~\cite{iso20059} measures the proportion of morphed samples incorrectly classified as bona fide. For face demorphing, an error occurs when the reconstruction retains high similarity to the trusted reference, exceeding $\tau$:
\begin{equation}
    \text{MACER}(\tau) = \frac{1}{N_M} \sum_{i=1}^{N_M} \mathbb{I}\left( S(I_{\text{out}}^{(i)}, I_{\text{ref}}^{(i)}) \ge \tau \right),
    \label{eq:macer}
\end{equation}
where $\mathbb{I}(\cdot)$ is the indicator function.

\textbf{Bona fide Sample Presentation Classification Error Rate (BSCER).}
BSCER~\cite{iso20059} measures the proportion of bona fide samples incorrectly classified as morphed samples. For face demorphing, an error occurs when the reconstruction degrades the identity of a bona fide document, resulting in a score below $\tau$:
\begin{equation}
    \text{BSCER}(\tau) = \frac{1}{N_{B}} \sum_{j=1}^{N_{B}} \mathbb{I}\left( S(I_{\text{out}}^{(j)}, I_{\text{ref}}^{(j)}) < \tau \right).
    \label{eq:bscer}
\end{equation}

\textbf{Equal Error Rate (EER).} EER~\cite{iso20059} is the operating point where MACER and BSCER are equal, providing a scalar summary of performance.

\subsubsection{Distributional Separability}
\textbf{Bona Fide-Morph Separability (BMS).} 
While D-MAD metrics evaluate performance at specific thresholds, they do not quantify the margin between bona fide and morphed classes. To evaluate fundamental separability, we introduce the BMS metric. We define score distributions for bona fide and morphed scenarios as:
\begin{equation}
\begin{aligned}
    \mathcal{D}_{\text{B}} &= \{ S(I_{\text{out}}^{(j)}, I_{\text{ref}}^{(j)}) \}_{j=1}^{N_{B}}, \\
    \mathcal{D}_{\text{M}} &= \{ S(I_{\text{out}}^{(i)}, I_{\text{ref}}^{(i)}) \}_{i=1}^{N_{M}}.
\end{aligned}
\label{eq:dmad_distributions}
\end{equation}

We compute the BMS as the 1-Wasserstein distance ($W_1$)~\cite{wasserstein_dist} between these distributions:
\begin{equation}
    \text{BMS} = W_1(\mathcal{D}_{\text{B}}, \mathcal{D}_{\text{M}}).
    \label{eq:bms}
\end{equation}
A higher BMS indicates a larger margin between bona fide and morphed scores, signifying a robust system less sensitive to threshold selection.

\subsection{Experimental Protocol} \label{sec:experimental_protocol}
We benchmark SFDemorpher against SOTA open-source face demorphing methods: Face Demorphing (FaDe)~\cite{face_demorphing}, StyleDemorpher-U (SD-U)~\cite{styledemorpher}, and StyleDemorpher-S (SD-S)~\cite{styledemorpher}. We refer to our method as \textbf{SFD}. As detailed in~\cite{styledemorpher}, SD-U was trained exclusively on landmark-based morphs, whereas SD-S was trained solely on deep learning-based morphs. For FaDe, we apply the recommended demorphing factor of $\bar{a}=0.3$. For all methods, we use the AdaFace~\cite{adaface} FRS for D-MAD (Fig.~\ref{fig:dmad_setup}). For the DTI and DNTI (Eqs.~\eqref{eq:DTI} and \eqref{eq:DNTI}), relying on a fixed decision threshold, we set $\tau=0.331$, corresponding to a $100\%$ True Match Rate (TMR) and $0.01\%$ False Match Rate (FMR) based on 96,214 mated and 41,404,391 non-mated identity pairs from DemorphDB~\cite{styledemorpher}.

\begin{figure*}[htbp]
    \centering
    \includegraphics[width=0.99\linewidth]{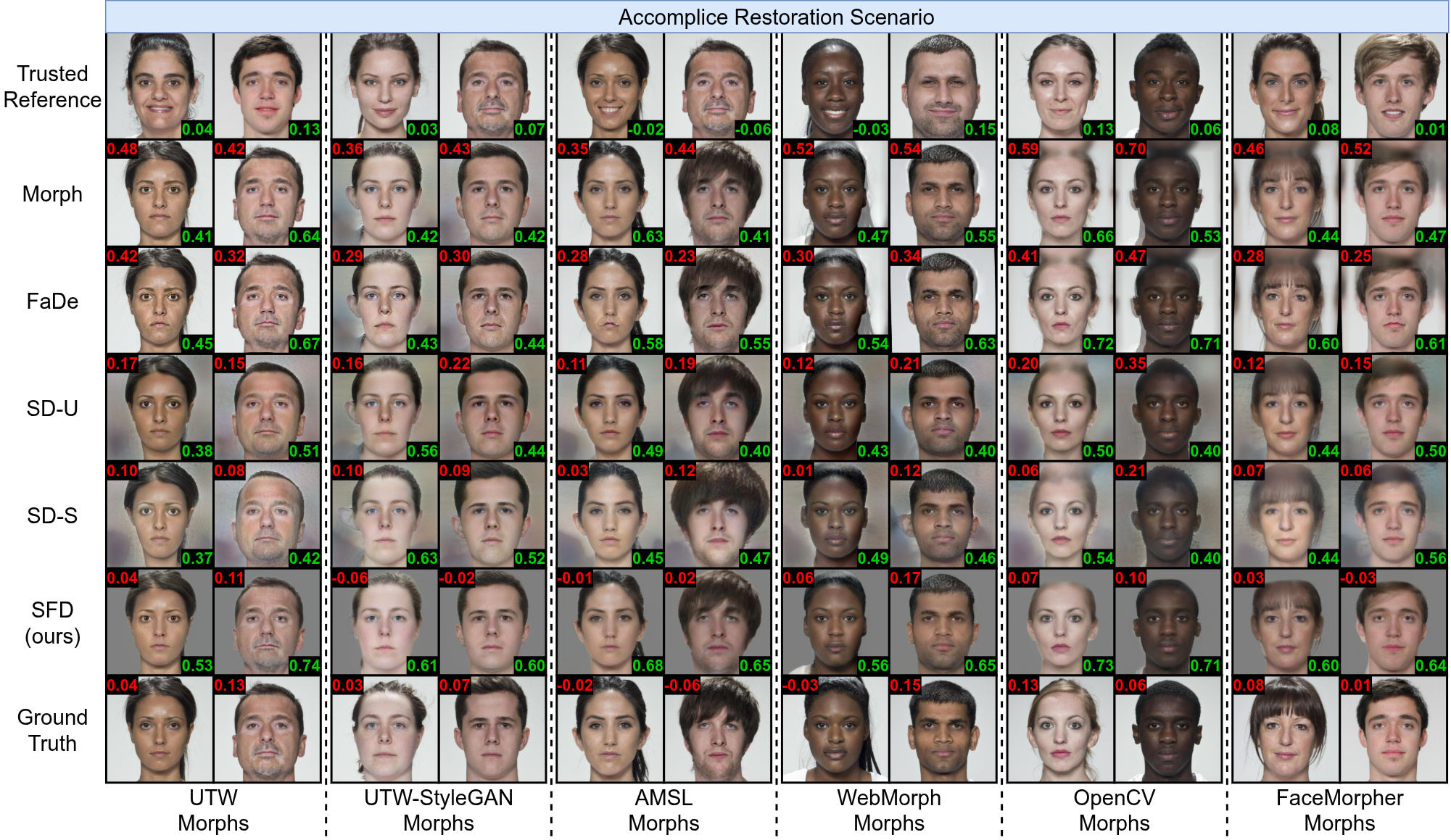}
    \caption{Qualitative results of accomplice identity restoration on morphed images from the FRLL-Morphs-UTW~\cite{frll_morphs_1, frll_morphs_2, styledemorpher} dataset. Green scores (bottom right) denote the identity similarity to the ground truth accomplice, while red scores (top left) denote similarity to the non-target criminal identity in the trusted reference. Effective demorphing yields high green scores and low red scores, indicating successful target reconstruction and non-target removal.}
    \label{fig:frll_restoration_visual}
\end{figure*}

We further compare our method with traditional D-MAD methods: Siamese~\cite{siamese}, CIFAA~\cite{iciap2023}, and ACIdA~\cite{acida}. Since these methods do not generate demorphed images, the DTI, DNTI, and BMS metrics are not computed for them.

\subsection{Evaluation} \label{sec:evaluation}
\subsubsection{Qualitative Evaluation}
Figure~\ref{fig:frll_restoration_visual} depicts accomplice restoration on FRLL-Morphs-UTW~\cite{frll_morphs_1, frll_morphs_2, styledemorpher}. Across all morphing methods, SFDemorpher accurately reconstructs the accomplice while suppressing the criminal identity. Conversely, SD-U~\cite{styledemorpher} and SD-S~\cite{styledemorpher} struggle to reconstruct the accomplice and fail to completely remove criminal traces, while FaDe~\cite{face_demorphing} often retains significant criminal features.

\begin{figure*}[htbp]
    \centering
    \includegraphics[width=0.99\linewidth]{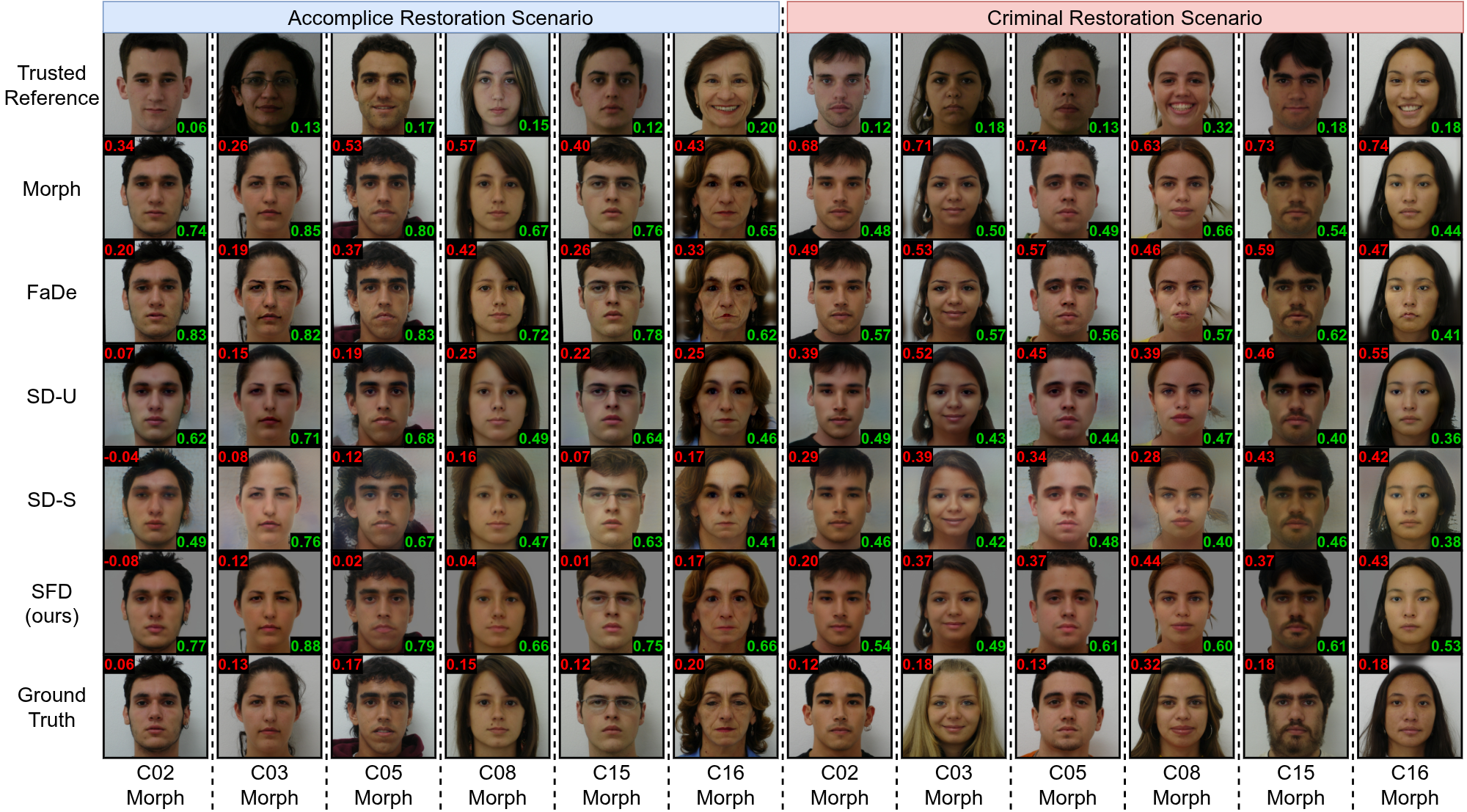}
    \caption{Qualitative results on the FEI Morph V2~\cite{iciap2023, feimorph} dataset on splicing-based~\cite{splicing} morphs. The visualization follows the same convention as Fig.~\ref{fig:frll_restoration_visual}. This dataset enables evaluation of both accomplice and criminal restoration scenarios.}
    \label{fig:fei_restoration_visual}
\end{figure*}

Figure~\ref{fig:fei_restoration_visual} visualizes accomplice and criminal restoration scenarios on the FEI Morph V2~\cite{iciap2023, feimorph} dataset. While accomplice restoration aligns with previous results, criminal restoration degrades across all methods. As discussed in Sec.~\ref{sec:problem_formulation}, this scenario is ill-posed due to missing outer facial data. Nevertheless, SFDemorpher achieves the best balance between accomplice removal and criminal reconstruction.

\begin{figure}[htbp]
    \centering
    \includegraphics[width=0.99\linewidth]{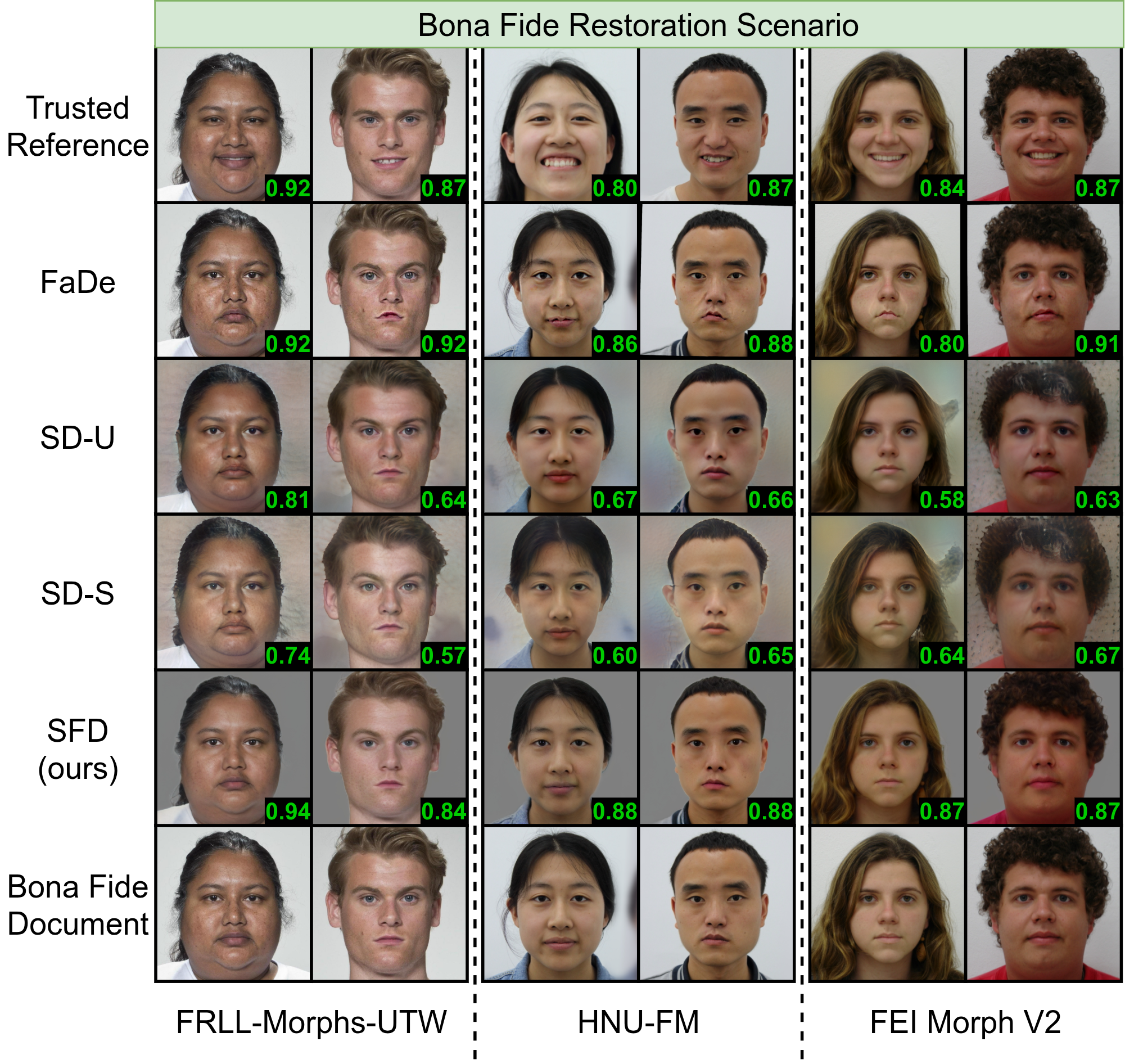}
    \caption{Qualitative results of bona fide identity restoration. The figure displays document-reference pairs from the FRLL-Morphs-UTW~\cite{frll_morphs_1, frll_morphs_2, styledemorpher}, HNU-FM~\cite{hnu_fm}, and FEI Morph V2~\cite{iciap2023, feimorph}. Green scores (bottom right) indicate the identity similarity to the bona fide document, where higher values signify superior identity preservation.}
    \label{fig:bonafide_restoration_visual}
\end{figure}

Figure~\ref{fig:bonafide_restoration_visual} visualizes demorphed outputs for the bona fide restoration scenario. SFDemorpher preserves identity fidelity with minimal degradation, achieving similarity scores comparable to trusted references. This contrasts with SD-U and SD-S, which degrade identity due to morph-only training. While FaDe preserves the identity information, it frequently introduces visual artifacts.

\subsubsection{Identity Deviation Analysis}
\begin{table*}[htbp]
\caption{Comparison of Deviation of Target (DTI) and Non-Target (DNTI) Identity for accomplice $(A)$ restoration on the HNU-FM~\cite{hnu_fm} and FRLL-Morphs-UTW~\cite{frll_morphs_1, frll_morphs_2, styledemorpher} datasets. Higher DTI$(A)$ and lower DNTI$(A)$ values indicate superior face demorphing performance.}
\label{tab:hnu_and_frll_deviations}
\centering
\footnotesize
\begin{tabular*}{\textwidth}{@{\extracolsep\fill}cccc}
\toprule
\multirow{2}{*}{\textbf{Dataset}} & \textbf{Demorphing} & \multirow{2}{*}{\textbf{DTI$(A)\uparrow$}} & \multirow{2}{*}{\textbf{DNTI$(A)\downarrow$}} \\
 & \textbf{Method} & & \\
\midrule
\multirow{5}{*}{HNU-FM}
& No Demorphing & 0.228 & 0.182 \\
& FaDe & \underline{0.311} & -0.004 \\
& SD-U & 0.170 & -0.125 \\
& SD-S & 0.190 & \textbf{-0.227} \\
& \textbf{SFD} (ours) & \textbf{0.335} & \underline{-0.221} \\
\midrule
\multirow{5}{*}{\shortstack{FRLL-Morphs-UTW \\ UTW \\ Morphs}}
& No Demorphing & 0.232 & 0.059 \\
& FaDe & \underline{0.249} & -0.088 \\
& SD-U & 0.154 & -0.190 \\
& SD-S & 0.097 & \underline{-0.274} \\
& \textbf{SFD} (ours) & \textbf{0.288} & \textbf{-0.317} \\
\midrule
\multirow{5}{*}{\shortstack{FRLL-Morphs-UTW \\ UTW-StyleGAN \\ Morphs}}
& No Demorphing & 0.094 & 0.069 \\
& FaDe & 0.112 & -0.110 \\
& SD-U & 0.103 & -0.140 \\
& SD-S & \underline{0.174} & \underline{-0.239} \\
& \textbf{SFD} (ours) & \textbf{0.220} & \textbf{-0.400} \\
\midrule
\multirow{5}{*}{\shortstack{FRLL-Morphs-UTW \\ FaceMorpher \\ Morphs}}
& No Demorphing & 0.224 & 0.170 \\
& FaDe & \underline{0.287} & -0.011 \\
& SD-U & 0.134 & -0.158 \\
& SD-S & 0.153 & \underline{-0.243} \\
& \textbf{SFD} (ours) & \textbf{0.344} & \textbf{-0.263} \\
\midrule
\multirow{5}{*}{\shortstack{FRLL-Morphs-UTW \\ WebMorph \\ Morphs}}
& No Demorphing & 0.233 & 0.176 \\
& FaDe & \underline{0.281} & 0.001 \\
& SD-U & 0.149 & -0.138 \\
& SD-S & 0.188 & \underline{-0.239} \\
& \textbf{SFD} (ours) & \textbf{0.350} & \textbf{-0.245} \\
\midrule
\multirow{5}{*}{\shortstack{FRLL-Morphs-UTW \\ AMSL \\ Morphs}}
& No Demorphing & 0.309 & 0.096 \\
& FaDe & \underline{0.365} & -0.073 \\
& SD-U & 0.203 & -0.190 \\
& SD-S & 0.205 & \underline{-0.269} \\
& \textbf{SFD} (ours) & \textbf{0.406} & \textbf{-0.328} \\
\midrule
\multirow{5}{*}{\shortstack{FRLL-Morphs-UTW \\ OpenCV \\ Morphs}}
& No Demorphing & 0.237 & 0.180 \\
& FaDe & \underline{0.302} & 0.001 \\
& SD-U & 0.142 & -0.149 \\
& SD-S & 0.165 & \underline{-0.235} \\
& \textbf{SFD} (ours) & \textbf{0.361} & \textbf{-0.241} \\
\bottomrule
\end{tabular*} 
\begin{flushleft}
\small \textbf{Bold} and \underline{underlined} values indicate the best and second-best results, respectively.
\end{flushleft}
\end{table*}

\begin{table*}[htbp]
\caption{Comparison of Deviation of Target Identity (DTI) and Non-Target Identity (DNTI) for accomplice $(A)$ and criminal $(C)$ restoration scenarios on the FEI Morph V2~\cite{iciap2023, feimorph} dataset.}
\label{tab:fei_deviations}
\centering
\footnotesize
\begin{tabular*}{\textwidth}{@{\extracolsep\fill}cccccc}
\toprule
\multirow{2}{*}{\textbf{Dataset}} & \textbf{Demorphing} & \multirow{2}{*}{\textbf{DTI$(A)\uparrow$}} & \multirow{2}{*}{\textbf{DNTI$(A)\downarrow$}} & \multirow{2}{*}{\textbf{DTI$(C)\uparrow$}} & \multirow{2}{*}{\textbf{DNTI$(C)\downarrow$}} \\
 & \textbf{Method} & & & & \\
\midrule
\multirow{5}{*}{\shortstack{FEI Morph V2 \\ C02 Morphs}}
& No Demorphing & 0.431 & 0.119 & 0.171 & 0.353 \\
& FaDe & \textbf{0.449} & -0.036 & \underline{0.200} & 0.198 \\
& SD-U & 0.276 & -0.131 & 0.088 & \underline{0.048} \\
& SD-S & 0.241 & \underline{-0.206} & 0.090 & \textbf{-0.030} \\
& \textbf{SFD} (ours) & \underline{0.432} & \textbf{-0.212} & \textbf{0.207} & 0.052 \\
\midrule
\multirow{5}{*}{\shortstack{FEI Morph V2 \\ C03 Morphs}}
& No Demorphing & \textbf{0.501} & 0.040 & 0.081 & 0.413 \\
& FaDe & \underline{0.487} & -0.110 & \underline{0.099} & 0.275 \\
& SD-U & 0.305 & -0.168 & 0.022 & \underline{0.110} \\
& SD-S & 0.260 & \underline{-0.246} & 0.022 & \textbf{0.031} \\
& \textbf{SFD} (ours) & 0.462 & \textbf{-0.250} & \textbf{0.111} & 0.152 \\
\midrule
\multirow{5}{*}{\shortstack{FEI Morph V2 \\ C05 Morphs}}
& No Demorphing & \textbf{0.484} & 0.066 & 0.111 & 0.399 \\
& FaDe & \underline{0.482} & -0.085 & \underline{0.133} & 0.256 \\
& SD-U & 0.299 & -0.152 & 0.044 & \underline{0.099} \\
& SD-S & 0.254 & \underline{-0.232} & 0.044 & \textbf{0.020} \\
& \textbf{SFD} (ours) & 0.455 & \textbf{-0.237} & \textbf{0.143} & 0.121 \\
\midrule
\multirow{5}{*}{\shortstack{FEI Morph V2 \\ C08 Morphs}}
& No Demorphing & \textbf{0.420} & 0.072 & 0.118 & 0.342 \\
& FaDe & 0.408 & -0.075 & \underline{0.130} & 0.201 \\
& SD-U & 0.253 & -0.148 & 0.041 & \underline{0.043} \\
& SD-S & 0.200 & \textbf{-0.224} & 0.031 & \textbf{-0.034} \\
& \textbf{SFD} (ours) & \underline{0.410} & \underline{-0.216} & \textbf{0.141} & 0.086 \\
\midrule
\multirow{5}{*}{\shortstack{FEI Morph V2 \\ C15 Morphs}}
& No Demorphing & \textbf{0.466} & 0.048 & 0.086 & 0.381 \\
& FaDe & \underline{0.449} & -0.099 & \underline{0.102} & 0.242 \\
& SD-U & 0.278 & -0.162 & 0.023 & \underline{0.080} \\
& SD-S & 0.230 & \textbf{-0.239} & 0.015 & \textbf{0.010} \\
& \textbf{SFD} (ours) & 0.440 & \underline{-0.237} & \textbf{0.112} & 0.114 \\
\midrule
\multirow{5}{*}{\shortstack{FEI Morph V2 \\ C16 Morphs}}
& No Demorphing & \textbf{0.381} & 0.090 & 0.139 & 0.308 \\
& FaDe & 0.367 & -0.058 & \underline{0.143} & 0.165 \\
& SD-U & 0.222 & -0.133 & 0.055 & \underline{0.015} \\
& SD-S & 0.170 & \textbf{-0.210} & 0.040 & \textbf{-0.063} \\
& \textbf{SFD} (ours) & \underline{0.374} & \underline{-0.195} & \textbf{0.152} & 0.051 \\
\bottomrule
\end{tabular*}
\begin{flushleft}
\small \textbf{Bold} and \underline{underlined} values indicate the best and second-best results, respectively.
\end{flushleft}
\end{table*}

\begin{table}[htbp]
\caption{Comparison of bona fide Deviation of Target Identity, DTI$(B)$, across face demorphing methods. Higher DTI$(B)$ values indicate better preservation of the original identity in demorphed bona fide documents.}
\label{tab:bonafide_deviation}
\centering
\footnotesize
\begin{tabular*}{\columnwidth}{@{\extracolsep\fill}ccc}
\toprule
\multirow{2}{*}{\textbf{Dataset}} & \textbf{Demorphing} & \multirow{2}{*}{\textbf{DTI$(B)\uparrow$}} \\
 & \textbf{Method} & \\
\midrule
\multirow{5}{*}{\shortstack{FRLL-Morphs- \\ UTW}}
& No Demorphing & \underline{0.569} \\
& FaDe & \textbf{0.597} \\
& SD-U & 0.327 \\
& SD-S & 0.308 \\
& \textbf{SFD} (ours) & 0.553 \\
\midrule
\multirow{5}{*}{HNU-FM}
& No Demorphing & \underline{0.572} \\
& FaDe & \textbf{0.616} \\
& SD-U & 0.366 \\
& SD-S & 0.340 \\
& \textbf{SFD} (ours) & 0.571 \\
\midrule
\multirow{5}{*}{FEI Morph V2}
& No Demorphing & 0.551 \\
& FaDe & \textbf{0.589} \\
& SD-U & 0.351 \\
& SD-S & 0.322 \\
& \textbf{SFD} (ours) & \underline{0.563} \\
\bottomrule
\end{tabular*}
\begin{flushleft}
\small \textbf{Bold} and \underline{underlined} values indicate the best and second-best results, respectively.
\end{flushleft}
\end{table}

Table~\ref{tab:hnu_and_frll_deviations} details accomplice reconstruction on HNU-FM and FRLL-Morphs-UTW. SFDemorpher consistently achieves the best DTI$(A)$ and DNTI$(A)$ values across all morphing methods. Notably, SFDemorpher improves upon the No~Demorphing scenario (where $I_{\hat{X}}$ is replaced by $I_{AC}$ in Eqs.~\eqref{eq:DTI} and \eqref{eq:DNTI}), indicating successful restoration of accomplice features suppressed in the morph. In contrast, SD-U and SD-S yield lower DTI$(A)$ scores. Regarding DNTI$(A)$, FaDe performs poorly by failing to remove the criminal's features. While SD-U and SD-S achieve improved DNTI$(A)$, SFDemorpher generally attains lower scores, showing effective target restoration and non-target suppression.

Table~\ref{tab:fei_deviations} summarizes FEI Morph V2 results. In the accomplice scenario, SFDemorpher shows marginally lower DTI$(A)$ than No~Demorphing and FaDe, suggesting minor identity information loss, yet it considerably outperforms SD-U and SD-S. For DNTI$(A)$, SFDemorpher performs similarly to SD-S, outperforming FaDe and SD-U. The slight performance decrease relative to FRLL-Morphs-UTW is likely due to the higher visual morph quality in FEI Morph V2 and greater variance in trusted reference images (see Fig.~\ref{fig:fei_restoration_visual}).

In the criminal restoration scenario, all methods perform worse on both DTI$(C)$ and DNTI$(C)$. Nevertheless, SFDemorpher achieves the highest DTI$(C)$, indicating superior restoration of missing criminal features. For DNTI$(C)$, SFDemorpher substantially outperforms FaDe; however, SD-U and SD-S achieve the lowest scores. This stems from their morph-only training, leading to aggressive removal of the identities in the trusted references. As SFDemorpher is trained on both bona fide and accomplice scenarios, the criminal scenario visually resembles bona fide training pairs due to the shared outer facial region. Consequently, SFDemorpher prioritizes identity preservation over aggressive disentanglement in this ill-posed setting, resulting in reduced non-target removal. Despite this, SFDemorpher still removes considerable accomplice features compared to the No~Demorphing baseline.

Table~\ref{tab:bonafide_deviation} evaluates bona fide restoration via the DTI$(B)$ metric. Ideally, processing a bona fide document should preserve identity without any modification. SFDemorpher achieves scores comparable to the No~Demorphing baseline across all datasets, indicating minimal degradation. Although FaDe yields slightly higher DTI$(B)$, SFDemorpher maintains identity fidelity equivalent to comparing bona fide documents to their trusted references. In contrast, SD-U and SD-S exhibit significantly lower DTI$(B)$. This degradation arises because both methods were trained exclusively on morphs, compromising bona fide performance.

Collectively, identity deviation results reveal a distinct trade-off among existing methods. FaDe demonstrates strong identity preservation (DTI) across bona fide and morphed scenarios but fails to meaningfully remove non-target features (DNTI). Conversely, SD-U and SD-S excel at aggressive non-target removal, particularly in the criminal scenario, but suffer from significant degradation in bona fide preservation and target restoration. SFDemorpher addresses this dichotomy by achieving a robust balance across all deviation metrics, which is critical for operational deployment. The impact of this balanced performance is further analyzed in Sec.~\ref{sec:DMAD_results} and \ref{sec:distributional_results}, where SFDemorpher's superiority in D-MAD and distributional separability is established.

\begin{table*}[htbp]
\caption{Comprehensive comparison of Differential Morphing Attack Detection (D-MAD) performance. The table reports Equal Error Rate (EER) and Bona fide Sample Presentation Classification Error Rate (BSCER) at fixed Morphing Attack Classification Error Rate (MACER) operating points~\cite{iso20059}.}
\label{tab:dmad_results}
\centering
\footnotesize
\begin{tabularx}{\textwidth}{@{\extracolsep\fill}cccYYY}
\toprule
\multirow{2}{*}{\textbf{Dataset}} & \multirow{2}{*}{\textbf{D-MAD Method}} & \multirow{2}{*}{\textbf{EER (\%)}} & \multicolumn{3}{c}{\textbf{BSCER @ MACER (\%)}} \\
\cmidrule{4-6}
 & & & \textbf{10\%} & \textbf{5\%} & \textbf{1\%} \\
\midrule
\multirow{8}{*}{FRLL-Morphs-UTW}
  & AdaFace & \textbf{0.08} & \textbf{0.00} & \textbf{0.00} & \textbf{0.00} \\
  & Siamese & 7.84 & 7.84 & 7.84 & 7.84 \\
  & CIFAA & 10.30 & 11.77 & 27.45 & 71.57 \\
  & ACIdA & 0.98 & \underline{0.98} & \underline{0.98} & \underline{0.98} \\
  & FaDe & \textbf{0.08} & \textbf{0.00} & \textbf{0.00} & \textbf{0.00} \\
  & SD-U & 0.23 & \textbf{0.00} & \textbf{0.00} & \textbf{0.00} \\
  & SD-S & \underline{0.09} & \textbf{0.00} & \textbf{0.00} & \textbf{0.00} \\
  & \textbf{SFD} (ours) & \textbf{0.08} & \textbf{0.00} & \textbf{0.00} & \textbf{0.00} \\
\midrule
\multirow{8}{*}{HNU-FM}
& AdaFace & 0.74 & 0.08 & 0.11 & 0.51 \\
 & Siamese & 4.88 & 2.86 & 4.81 & 9.73 \\
 & CIFAA & 5.98 & 3.52 & 6.97 & 13.18 \\
 & ACIdA & 2.07 & 0.28 & 0.81 & 4.05 \\
 & FaDe & 0.78 & 0.08 & 0.20 & 0.65 \\
 & SD-U & 1.34 & 0.06 & 0.17 & 1.78 \\
 & SD-S & \underline{0.65} & \underline{0.01} & \underline{0.10} & \underline{0.46} \\
 & \textbf{SFD} (ours) & \textbf{0.27} & \textbf{0.00} & \textbf{0.00} & \textbf{0.11} \\
\midrule
\multirow{8}{*}{\shortstack{FEI Morph V2 \\ Accomplice \\ Restoration}}
  & AdaFace & \textbf{0.20} & \textbf{0.00} & \textbf{0.00} & \textbf{0.00} \\
  & Siamese & 6.43 & 3.00 & 9.50 & 38.75 \\
  & CIFAA & 14.04 & 18.75 & 29.25 & 51.50 \\
  & ACIdA & 2.25 & \underline{0.50} & \underline{1.25} & 3.00 \\
  & FaDe & \underline{0.25} & \textbf{0.00} & \textbf{0.00} & \textbf{0.00} \\
  & SD-U & 0.50 & \textbf{0.00} & \textbf{0.00} & \underline{0.25} \\
  & SD-S & 0.50 & \textbf{0.00} & \textbf{0.00} & \textbf{0.00} \\
  & \textbf{SFD} (ours) & \underline{0.25} & \textbf{0.00} & \textbf{0.00} & \textbf{0.00} \\
\midrule
\multirow{8}{*}{\shortstack{FEI Morph V2 \\ Criminal \\ Restoration}}
  & AdaFace & 11.88 & 16.00 & 24.25 & 45.25 \\
  & Siamese & 15.82 & 23.00 & 34.75 & 57.50 \\
  & CIFAA & 14.26 & 19.00 & 30.00 & 52.75 \\
  & ACIdA & \underline{10.50} & \underline{11.00} & \underline{20.75} & \underline{33.75} \\
  & FaDe & 13.56 & 18.25 & 27.75 & 45.50 \\
  & SD-U & 12.44 & 16.25 & 29.50 & 57.50 \\
  & SD-S & 11.75 & 12.75 & 22.50 & 47.50 \\
  & \textbf{SFD} (ours) & \textbf{5.33} & \textbf{2.25} & \textbf{5.50} & \textbf{21.00} \\
\bottomrule
\end{tabularx}
\begin{flushleft}
\small \textbf{Bold} and \underline{underlined} values indicate the best and second-best results, respectively.
\end{flushleft}
\end{table*}
\subsubsection{Morphing Attack Detection Performance} \label{sec:DMAD_results}
\begin{figure*}[htbp]
    \centering
    \begin{subfigure}[htbp]{0.38\linewidth}
        \centering
        \includegraphics[width=\linewidth]{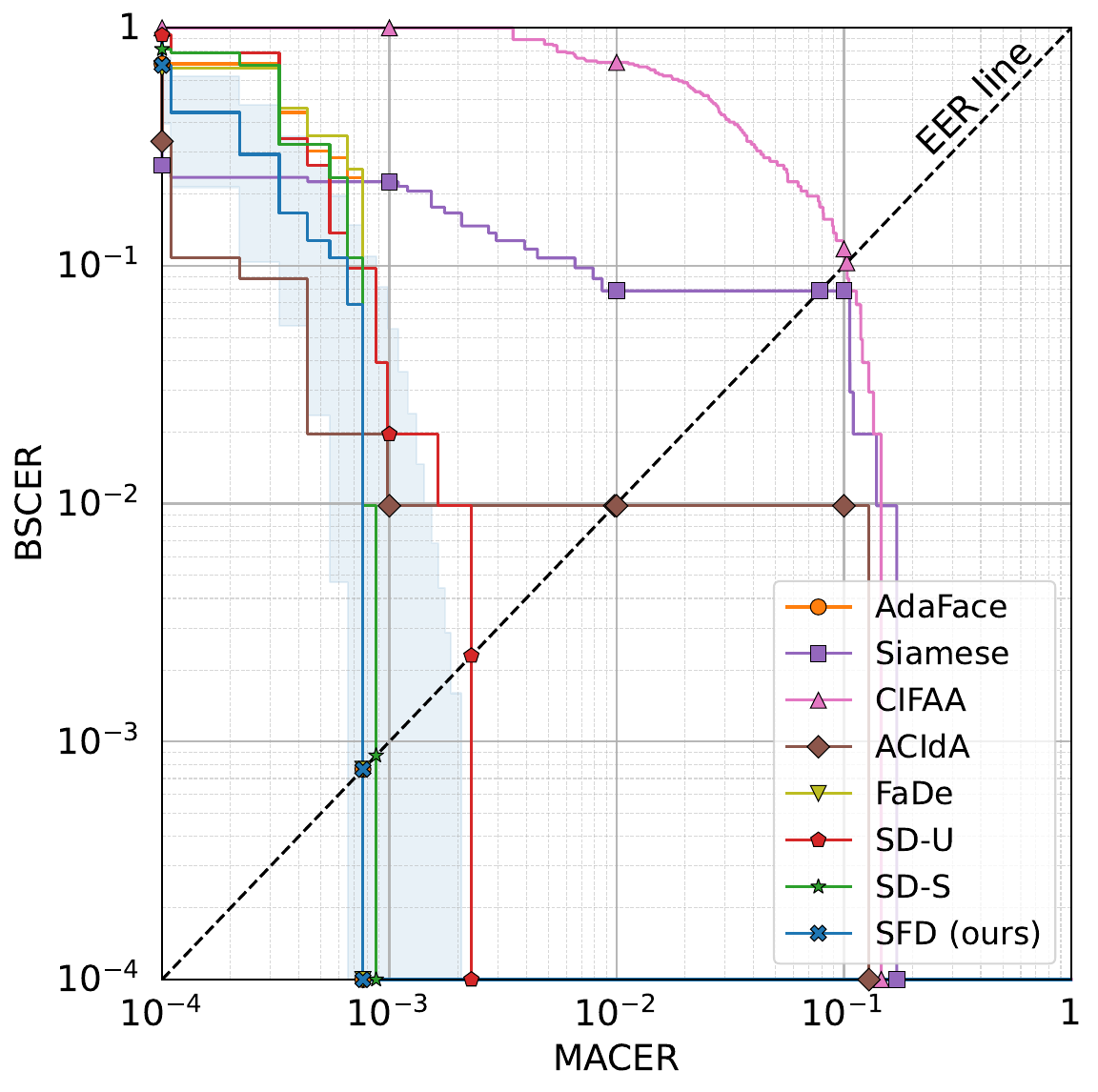}
        \caption{FRLL-Morphs-UTW}
        \label{fig:frll}
    \end{subfigure}\hspace{1cm}
    \begin{subfigure}[htbp]{0.38\linewidth}
        \centering
        \includegraphics[width=\linewidth]{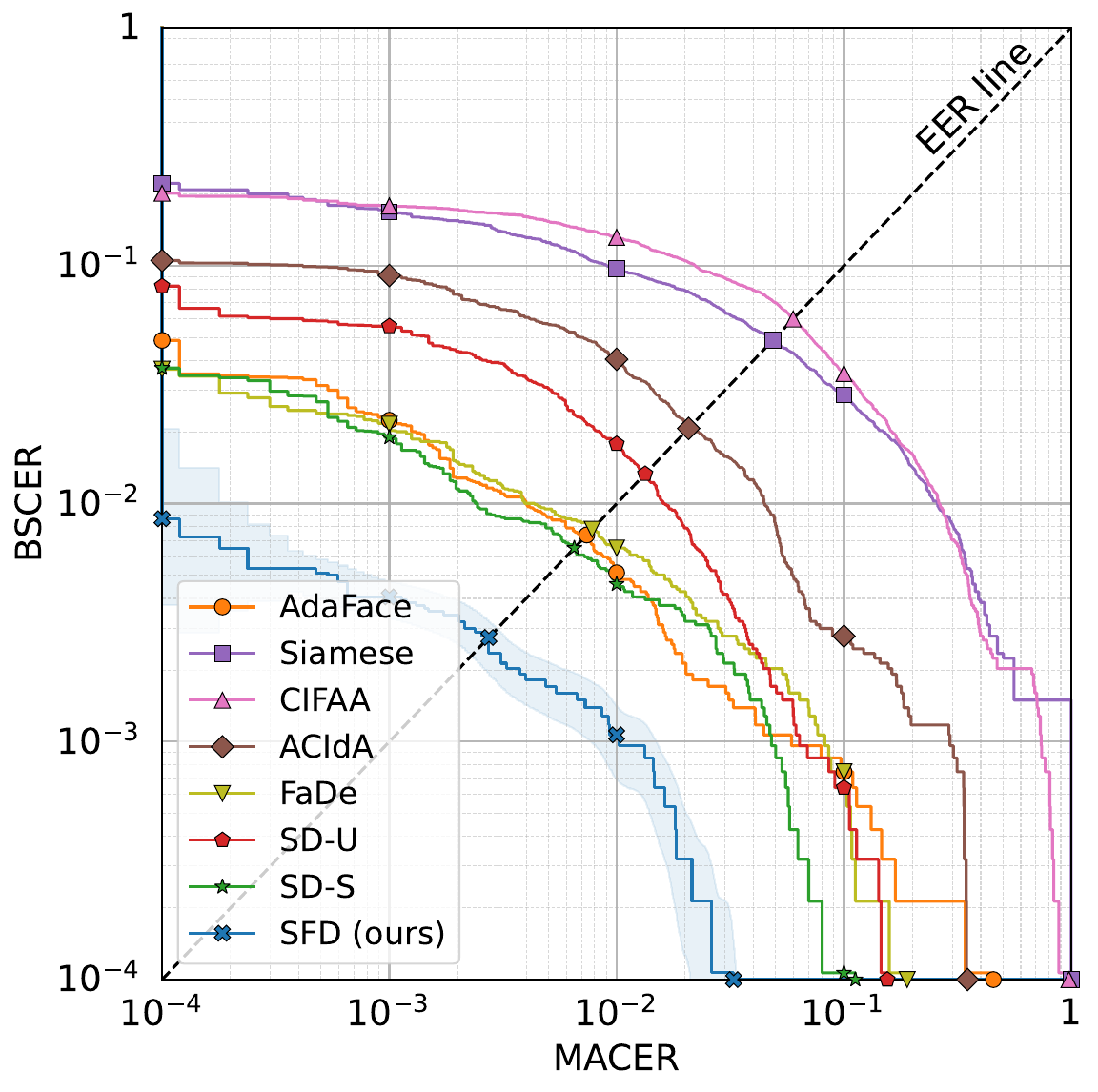}
        \caption{HNU-FM}
        \label{fig:hnu}
    \end{subfigure}
    \vspace{1em} 
    \begin{subfigure}[htbp]{0.38\linewidth}
        \centering
        \includegraphics[width=\linewidth]{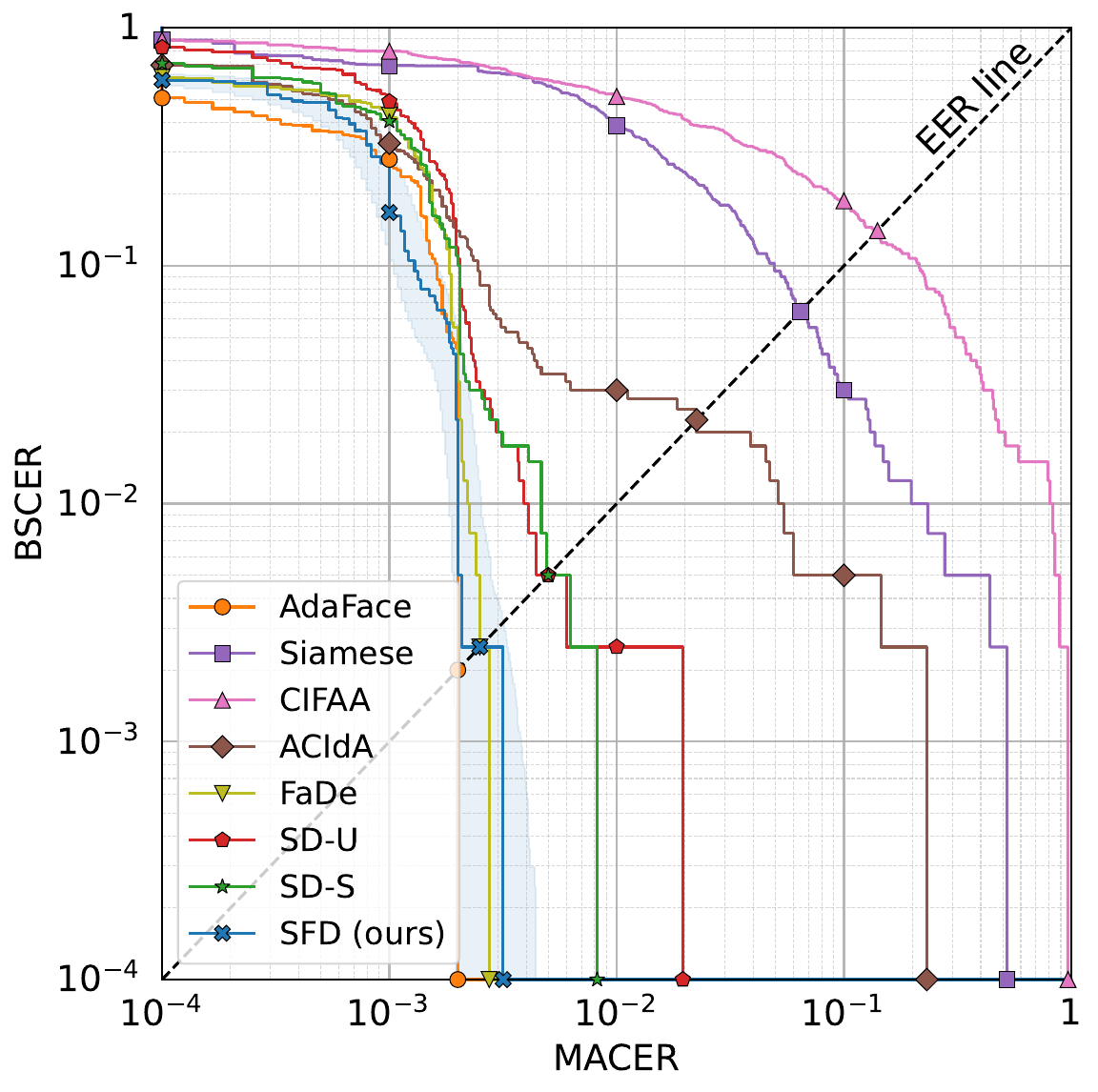}
        \caption{FEI Morph V2 Accomplice Restoration}
        \label{fig:fei_acc}
    \end{subfigure}\hspace{1cm}
    \begin{subfigure}[htbp]{0.38\linewidth}
        \includegraphics[width=\linewidth]{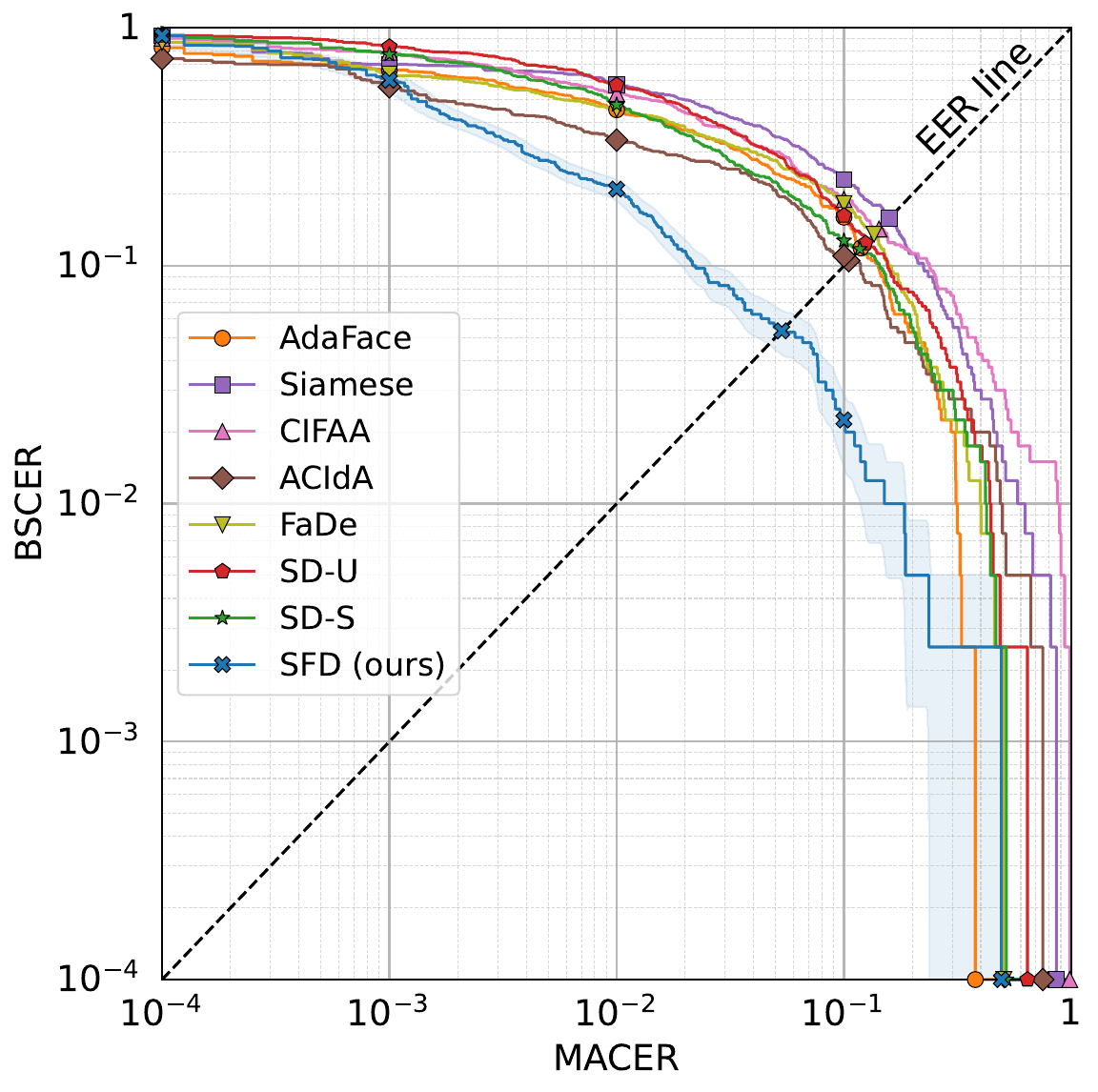} 
        \caption{FEI Morph V2 Criminal Restoration}
        \label{fig:fei_crim}
    \end{subfigure}
    \caption{Detection Error Trade-off (DET) curves showing D-MAD performance across evaluation datasets. For our proposed approach (SFD), the shaded region represents the $\pm1$ standard deviation interval.}
    \label{fig:det_curves}
\end{figure*}

Table~\ref{tab:dmad_results} presents the D-MAD performance across all evaluation datasets. Following the D-MAD pipeline established in Sec.~\ref{sec:dmad_integration}, we employ AdaFace~\cite{adaface} as the underlying FRS for all demorphing-based methods to ensure a fair comparison. Due to the large number of morphing methods, results are computed across all morphing methods per dataset, while the performance on each morphing method remains consistent with these aggregated trends. Notably, SFDemorpher achieves state-of-the-art results across all datasets.

In FRLL-Morphs-UTW and FEI Morph V2 accomplice scenarios, most demorphing methods achieve low Equal Error Rates (EER), performing comparably to the AdaFace baseline. This indicates that the underlying FRS is highly effective in these conditions, and provided the demorphing process does not significantly degrade image quality, detection performance remains high. In contrast, traditional D-MAD methods (Siamese~\cite{siamese}, CIFAA~\cite{iciap2023}, ACIdA~\cite{acida}) act as outliers, performing substantially worse than even the FRS baseline. This gap highlights a key architectural advantage: the demorphing pipeline is modular, allowing performance to scale directly with advancements in FRS technology. Traditional D-MAD methods, reliant on detecting specific morphing artifacts, cannot benefit from FRS improvements and often struggle with advanced morphing attacks.

The differentiation between methods is more pronounced in the HNU-FM and FEI Morph V2 criminal scenario. On HNU-FM, SFDemorpher achieves the lowest EER (0.27\%), outperforming the AdaFace baseline (0.74\%) and other demorphing methods. The most significant improvement is in the FEI Morph V2 criminal restoration scenario, which is inherently challenging due to missing facial information~\cite{acida}. Here, SFDemorpher reduces the EER to 5.33\%, outperforming the AdaFace baseline (11.88\%) and all other D-MAD methods. While accomplice restoration is largely driven by the strength of the FRS, the criminal scenario requires more nuanced identity disentanglement without compromising bona fide restoration performance. SFDemorpher's dual-pass training enables this equilibrium, minimizing errors on both morphed and bona fide images.

Figure~\ref{fig:det_curves} illustrates the Detection Error Trade-off (DET) curves, providing a visual confirmation of the observed trends. Across all datasets, traditional D-MAD methods exhibit the highest error rates. Demorphing-based methods generally cluster closer to the FRS baseline, but SFDemorpher consistently demonstrates superior performance. Results confirm that while a strong FRS is beneficial, our framework further enhances detection capability by rectifying identity inconsistencies in morphed documents. 

\subsubsection{Distributional Analysis} \label{sec:distributional_results}
Table~\ref{tab:bms_results} reports the Bona Fide-Morph Separability (BMS) scores, quantifying the margin between the bona fide ($\mathcal{D}_{\text{B}}$) and morphed ($\mathcal{D}_{\text{M}}$) score distributions defined in Eq.~\eqref{eq:dmad_distributions}. SFDemorpher achieves the highest BMS scores across all datasets, significantly outperforming other methods. Notably, even for FRLL-Morphs-UTW and FEI Morph V2 accomplice restoration datasets, where EER results are comparable due to the strong AdaFace FRS (Tab.~\ref{tab:dmad_results}), SFDemorpher demonstrates superior separability. This confirms that while error rates may converge when the FRS is highly effective, SFDemorpher establishes a wider margin between bona fide and morphed scores.

\begin{table}[htbp]
\caption{Comparison of Bona Fide-Morph Separability (BMS) scores. Higher BMS values indicate a larger margin between the score distributions of bona fide and morphed samples} 
\label{tab:bms_results}
\centering
\footnotesize
\begin{tabular*}{\columnwidth}{@{\extracolsep\fill}ccc}
\toprule
\multirow{2}{*}{\textbf{Dataset}} & \textbf{Demorphing} & \multirow{2}{*}{\textbf{BMS$\uparrow$}} \\
 & \textbf{Method} & \\
\midrule
\multirow{5}{*}{FRLL-Morphs-UTW}
& No Demorphing & 0.453 \\
& FaDe & \underline{0.551} \\
& SD-U & 0.427 \\
& SD-S & 0.489 \\
& \textbf{SFD} (ours) & \textbf{0.735} \\
\midrule
\multirow{5}{*}{HNU-FM}
& No Demorphing & 0.390 \\
& FaDe & \underline{0.508} \\
& SD-U & 0.410 \\
& SD-S & 0.478 \\
& \textbf{SFD} (ours) & \textbf{0.665} \\
\midrule
\multirow{5}{*}{\shortstack{FEI Morph V2 \\ Accomplice}}
& No Demorphing & 0.478 \\
& FaDe & \underline{0.551} \\
& SD-U & 0.423 \\
& SD-S & 0.466 \\
& \textbf{SFD} (ours) & \textbf{0.652} \\
\midrule
\multirow{5}{*}{\shortstack{FEI Morph V2 \\ Criminal}}
& No Demorphing & 0.185 \\
& FaDe & \underline{0.251} \\
& SD-U & 0.207 \\
& SD-S & \underline{0.251} \\
& \textbf{SFD} (ours) & \textbf{0.331} \\
\bottomrule
\end{tabular*}
\begin{flushleft}
\small \textbf{Bold} and \underline{underlined} values indicate the best and second-best results, respectively.
\end{flushleft}
\end{table}

\begin{figure*}[htbp]
    \centering
    \begin{subfigure}[htbp]{0.24\linewidth}
        \centering
        \includegraphics[width=0.99\linewidth]{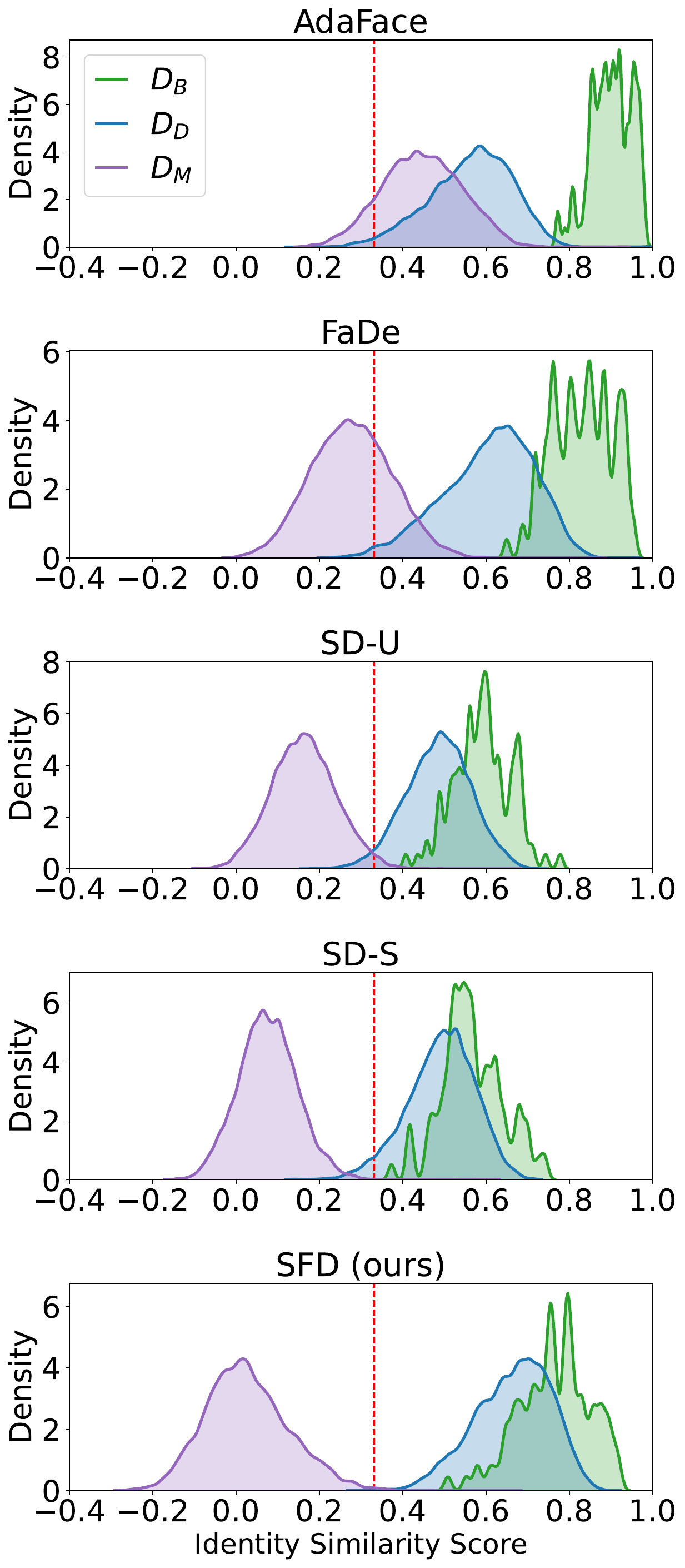}
        \caption{FRLL-Morphs-UTW}
        \label{fig:hist_frll}
    \end{subfigure}
    \hfill
    \begin{subfigure}[htbp]{0.24\linewidth}
        \centering
        \includegraphics[width=0.99\linewidth]{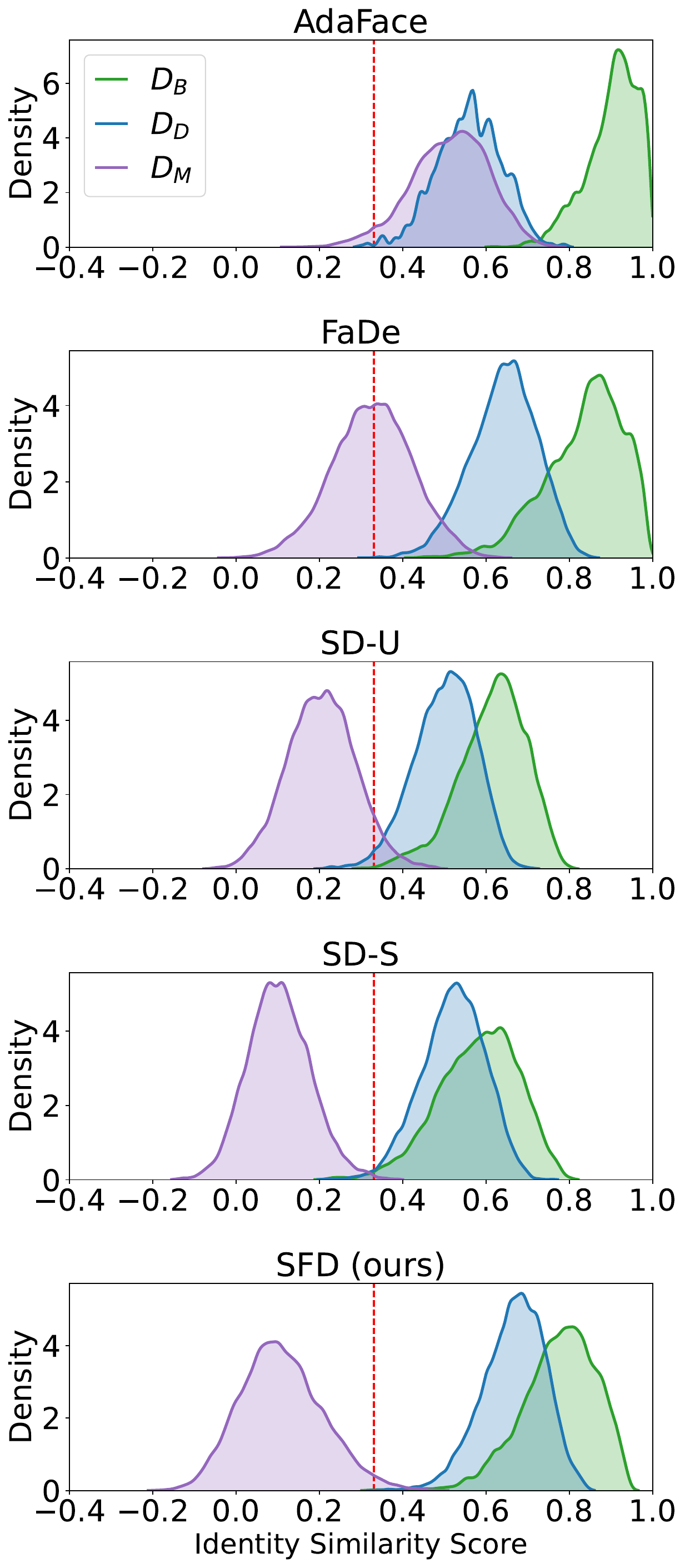}
        \caption{HNU-FM}
        \label{fig:hist_hnu}
    \end{subfigure}
    \hfill
    \begin{subfigure}[htbp]{0.24\linewidth}
        \centering
        \includegraphics[width=0.99\linewidth]{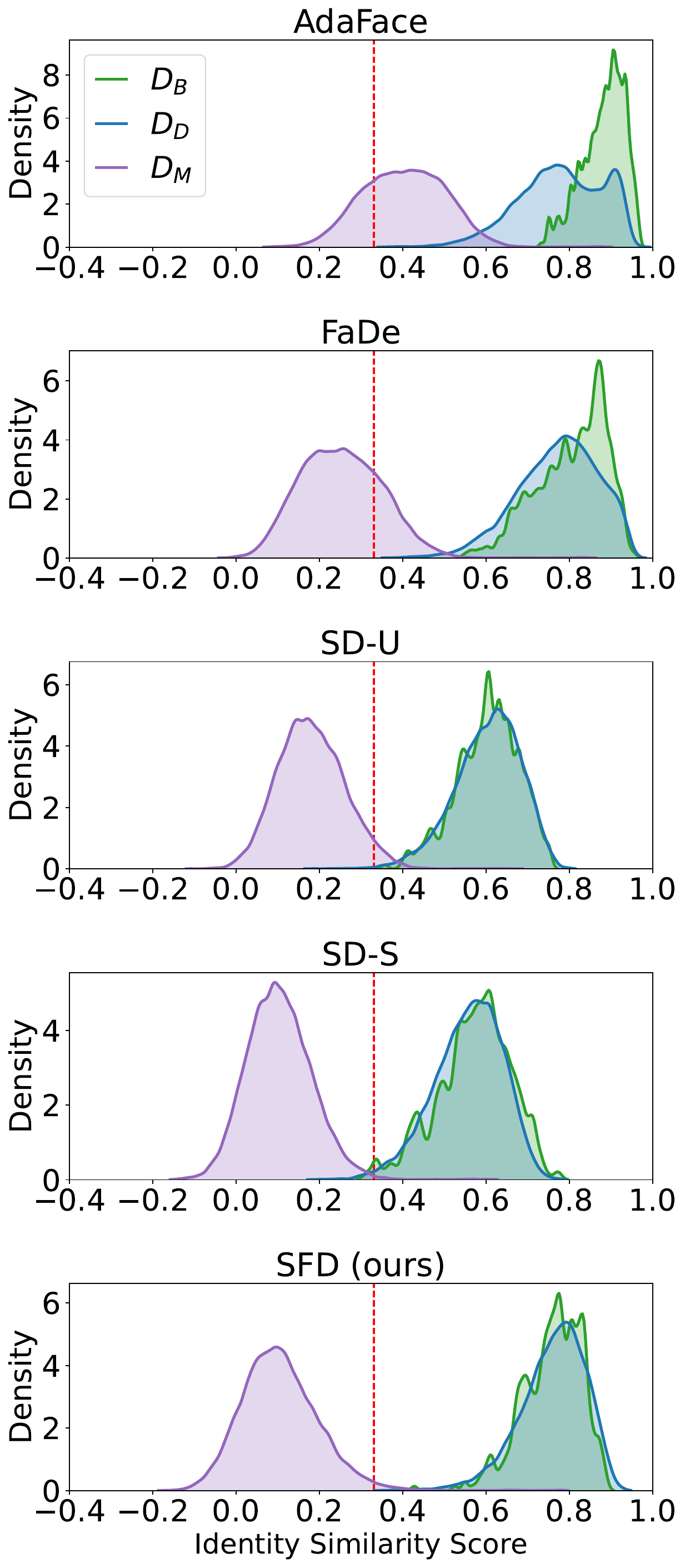}
        \caption{FEI Morph V2 Acc.}
        \label{fig:hist_fei_acc}
    \end{subfigure}
    \hfill
    \begin{subfigure}[htbp]{0.24\linewidth}
        \centering
        \includegraphics[width=0.99\linewidth]{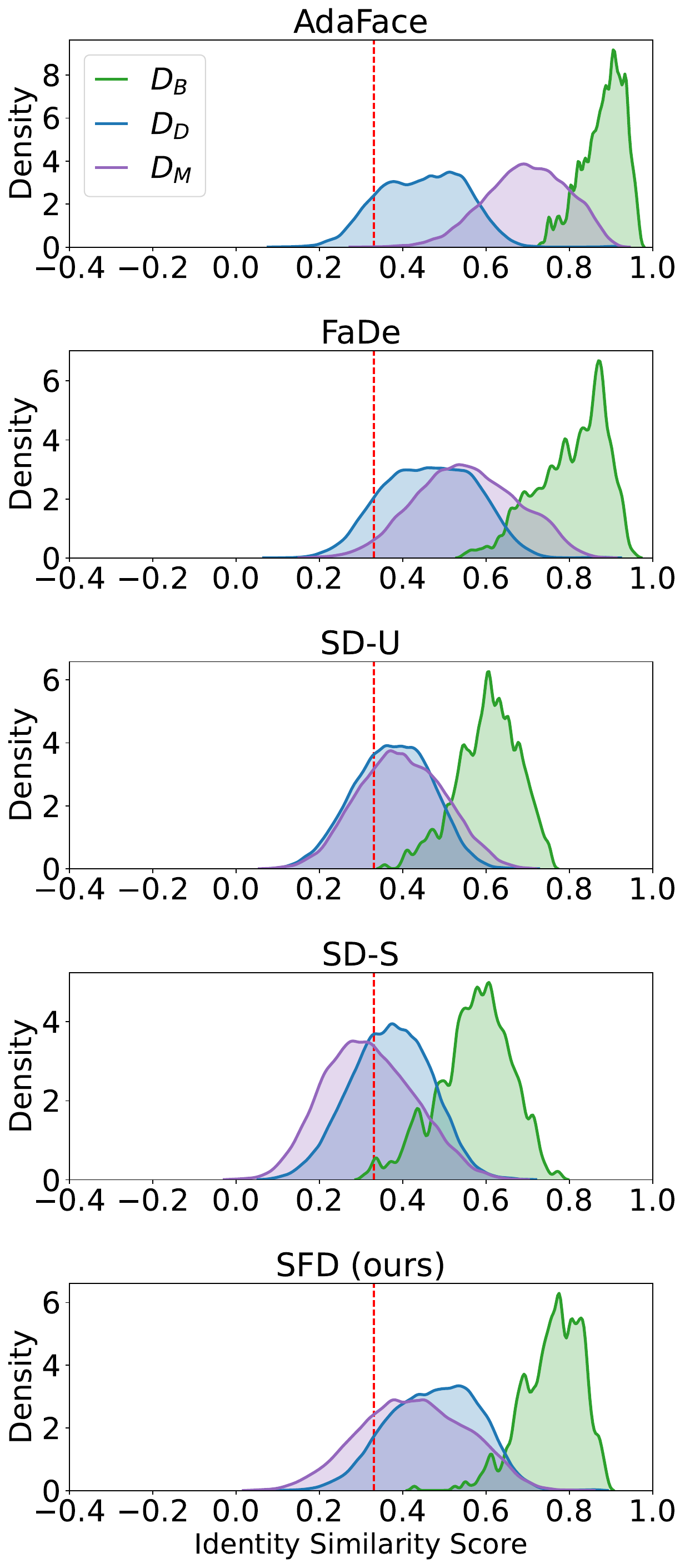}
        \caption{FEI Morph V2 Crim.}
        \label{fig:hist_fei_cri}
    \end{subfigure}
    \caption{Identity similarity score histograms of different evaluation datasets. Green ($\mathcal{D}_{\text{B}}$) and purple ($\mathcal{D}_{\text{M}}$) distributions represent bona fide and morphed score sets (Eq.~\eqref{eq:dmad_distributions}). The blue distribution $\mathcal{D}_{\text{D}}$ (Eq.~\eqref{eq:dd_distribution}) represents the similarity between the demorphed output and the ground truth target identity. The red dotted line denotes the FRS decision threshold $\tau$. Ideally, a robust method preserves $\mathcal{D}_{\text{B}}$ near the AdaFace baseline, shifts $\mathcal{D}_{\text{M}}$ left of $\tau$ for non-target removal, and positions $\mathcal{D}_{\text{D}}$ right of $\tau$ for target recovery.}
    \label{fig:score_histograms}
\end{figure*}

Figure~\ref{fig:score_histograms} visualizes these score distributions alongside a third distribution, $\mathcal{D}_{\text{D}}$, representing the similarity between the demorphed output and the ground truth target identity:
\begin{equation}
    \mathcal{D}_{\text{D}} = \{ S(I_{\text{out}}^{(i)}, I_{\text{GT}}^{(i)}) \}_{i=1}^{N_{M}}.
    \label{eq:dd_distribution}
\end{equation}
In these plots, $\mathcal{D}_{\text{B}}$ (green) and $\mathcal{D}_{\text{M}}$ (purple) determine D-MAD performance, where minimal overlap is ideal. The FRS decision threshold $\tau$ (red dotted line) separates the classes. Additionally, $\mathcal{D}_{\text{D}}$ (blue) validates reconstruction quality; ideally, it should reside above $\tau$, indicating successful target identity recovery. For AdaFace, $I_{\text{out}}$ is replaced by the input document image depending on the scenario ($I_{B}$ or $I_{AC}$).

Analyzing $\mathcal{D}_{\text{B}}$, SFDemorpher and FaDe maintain distributions close to the AdaFace baseline, indicating minimal impact on bona fide images. In contrast, SD-U and SD-S exhibit a significant leftward shift, reflecting the identity degradation observed in DTI$(B)$ results. For $\mathcal{D}_{\text{M}}$, FaDe struggles to shift the distribution below $\tau$, aligning with its poor DNTI performance where non-target identity features persist. SD-U and SD-S achieve better separation but often at the cost of target identity loss. SFDemorpher achieves the most effective shift of $\mathcal{D}_{\text{M}}$ to the left while preserving $\mathcal{D}_{\text{B}}$, resulting in the minimal overlap. 

For the $\mathcal{D}_{\text{D}}$ distributions, SFDemorpher and FaDe show a rightward shift relative to the AdaFace baseline, confirming that target identity information is enhanced in the output. Conversely, SD-U and SD-S exhibit a leftward shift, indicating that target identity features are lost during the aggressive disentanglement process. 

In the challenging FEI Morph V2 criminal restoration scenario, all methods exhibit greater overlap between $\mathcal{D}_{\text{B}}$ and $\mathcal{D}_{\text{M}}$, with $\mathcal{D}_{\text{M}}$ struggling to move below $\tau$ due to the ill-posed nature of the task. Nevertheless, SFDemorpher maintains the smallest overlap and largest separation among all methods. This distributional analysis corroborates all previous results: SFDemorpher's ability to maximize the margin between $\mathcal{D}_{\text{B}}$ and $\mathcal{D}_{\text{M}}$ while ensuring $\mathcal{D}_{\text{D}}$ remains high validates its robustness and operational viability.

\subsubsection{Ablation Studies}

\begin{figure*}[htbp]
    \centering
    \begin{subfigure}[htbp]{0.32\linewidth}
        \centering
        \includegraphics[width=\linewidth]{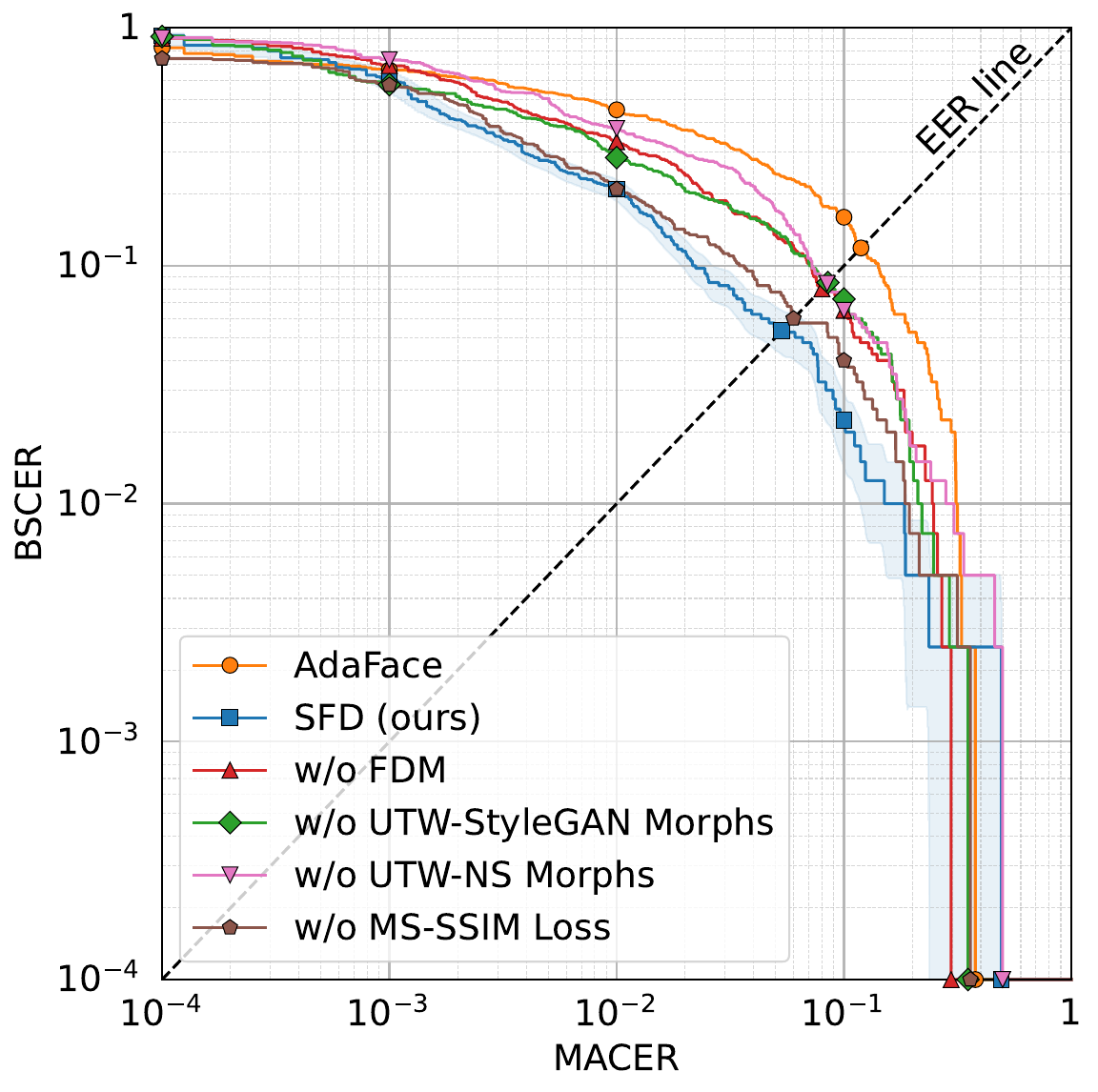}
        \caption{Training components}
        \label{fig:ablation_changes}
    \end{subfigure}\hfill
    \begin{subfigure}[htbp]{0.32\linewidth}
        \centering
        \includegraphics[width=\linewidth]{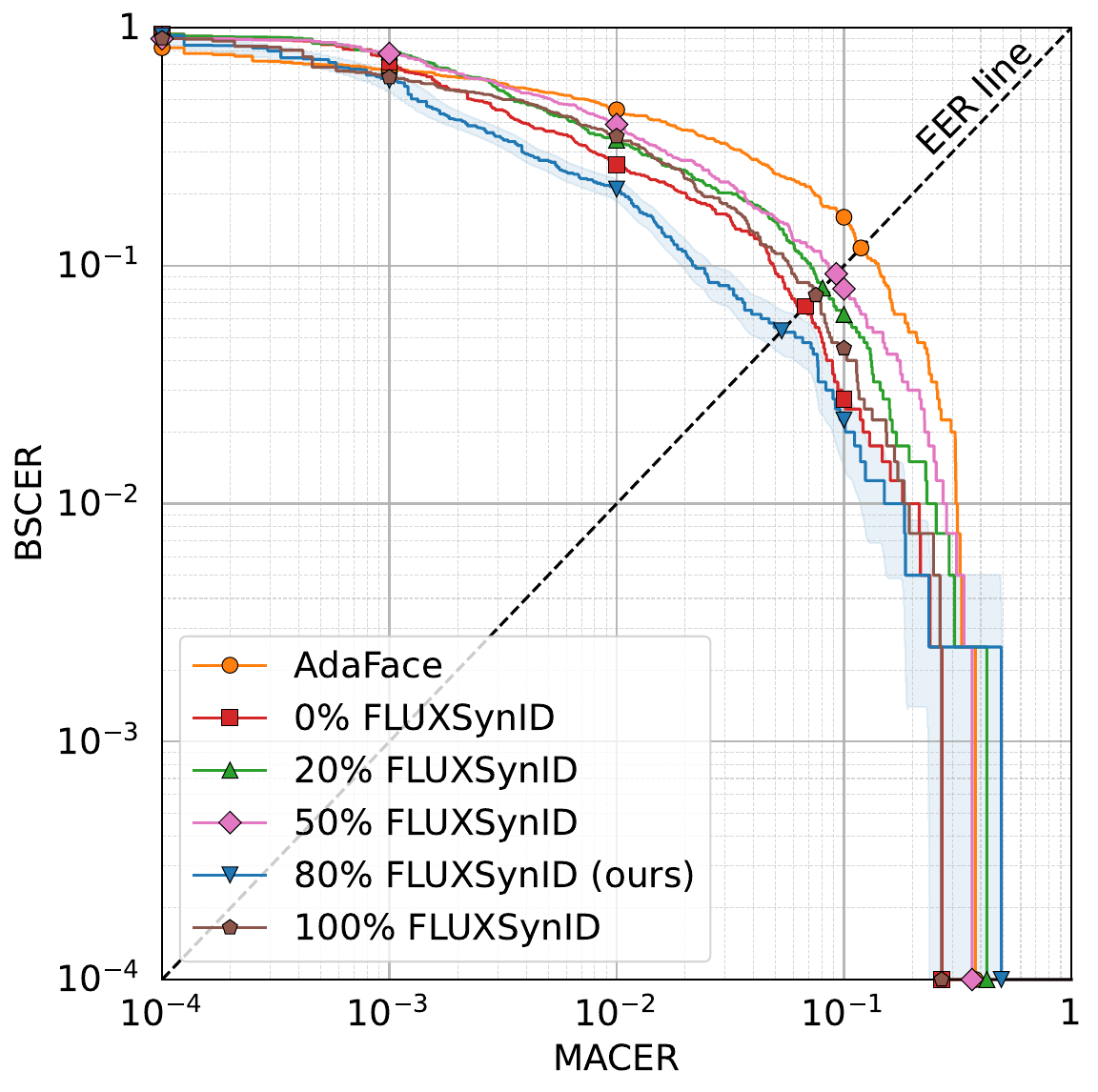}
        \caption{Synthetic data ratios}
        \label{fig:ablation_syn_data}
    \end{subfigure}\hfill
    \begin{subfigure}[htbp]{0.32\linewidth}
        \centering
        \includegraphics[width=\linewidth]{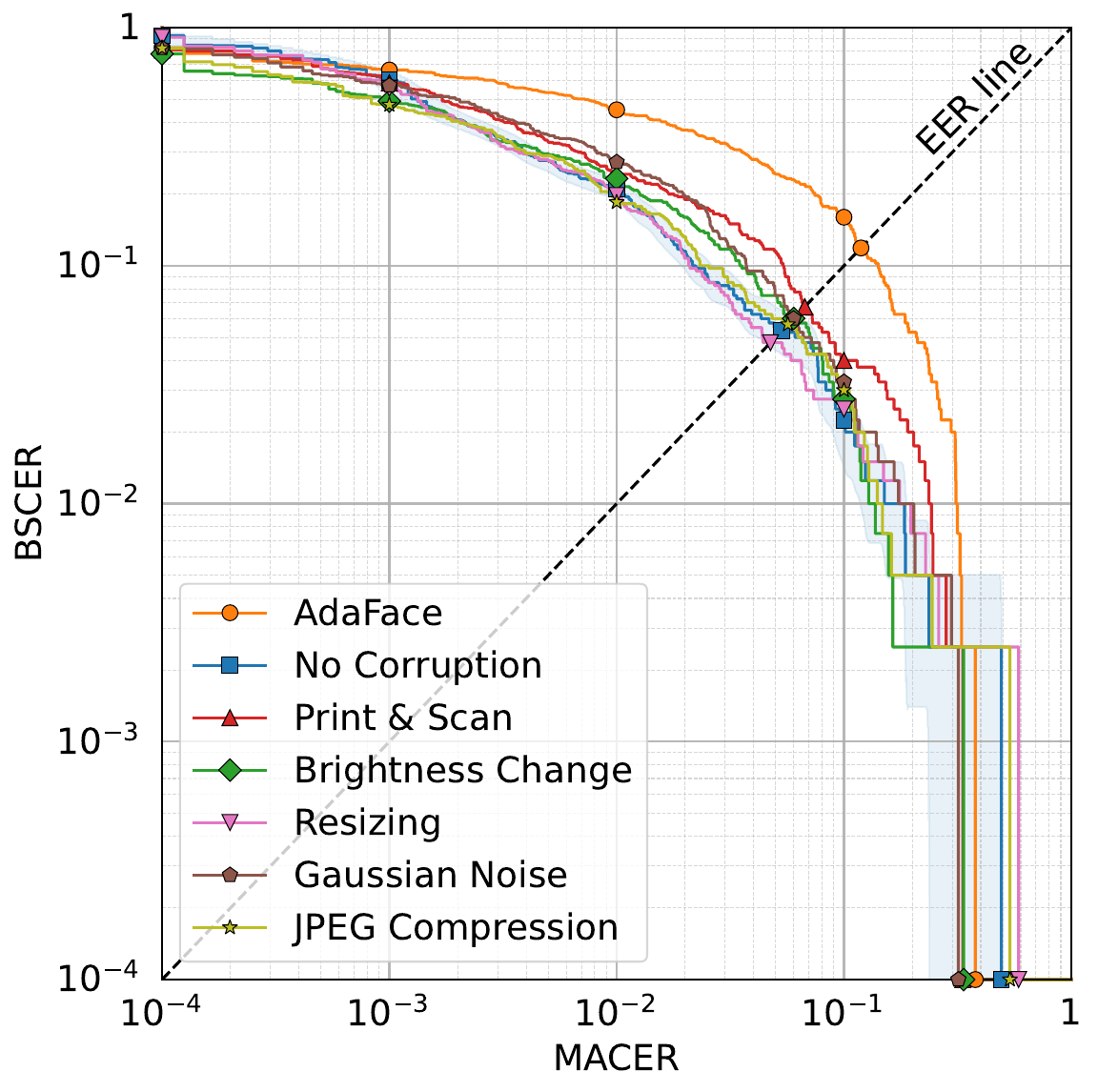}
        \caption{Robustness to corruptions}
        \label{fig:ablation_corruptions}
    \end{subfigure}
        \\ \vspace{1em}
    \begin{subfigure}[htbp]{0.32\linewidth}
        \centering
        \includegraphics[width=\linewidth]{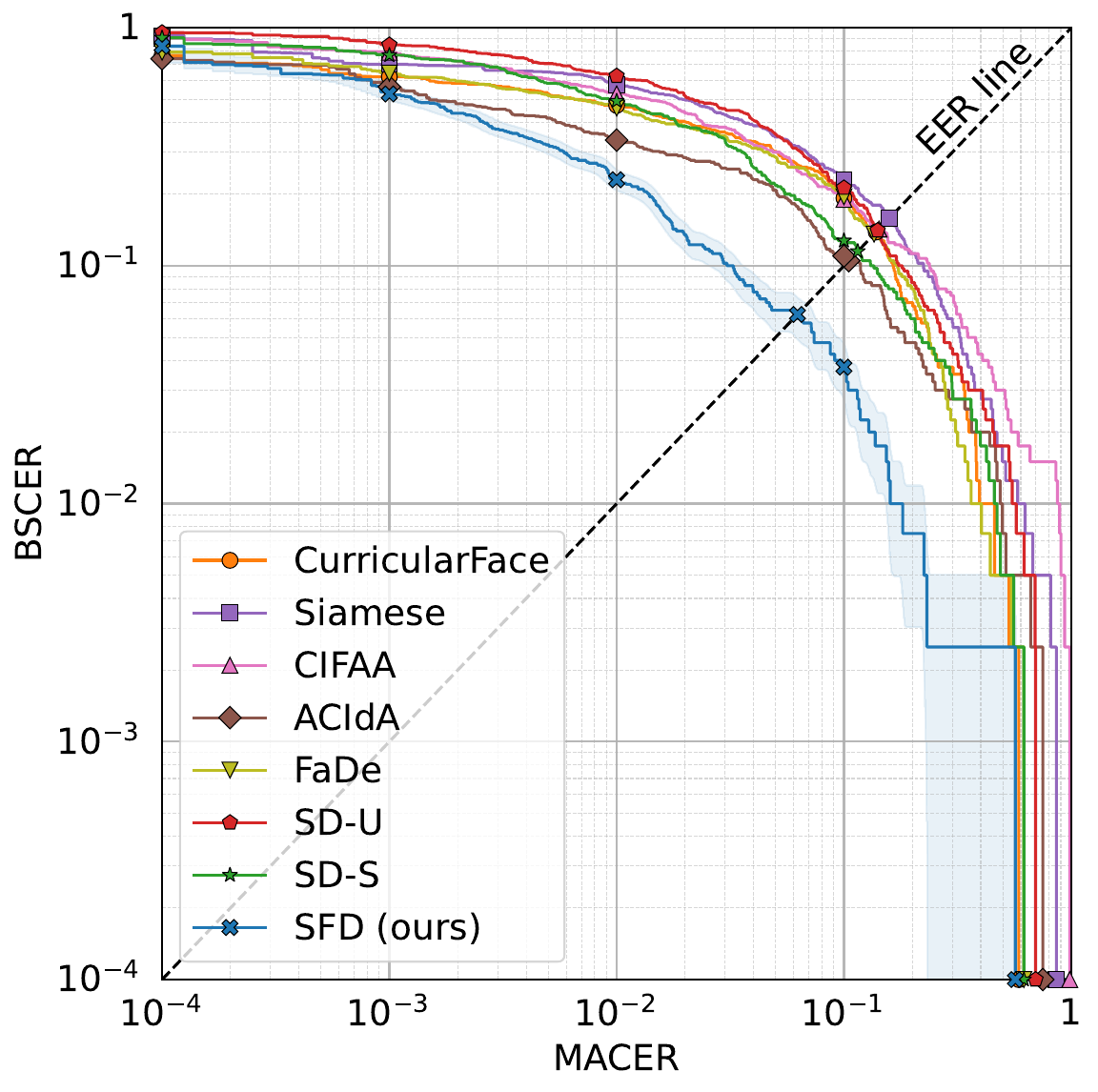}
        \caption{CurricularFace FRS}
        \label{fig:ablation_curricularface}
    \end{subfigure}\hspace{0.5cm}
    \begin{subfigure}[htbp]{0.32\linewidth}
        \centering
        \includegraphics[width=\linewidth]{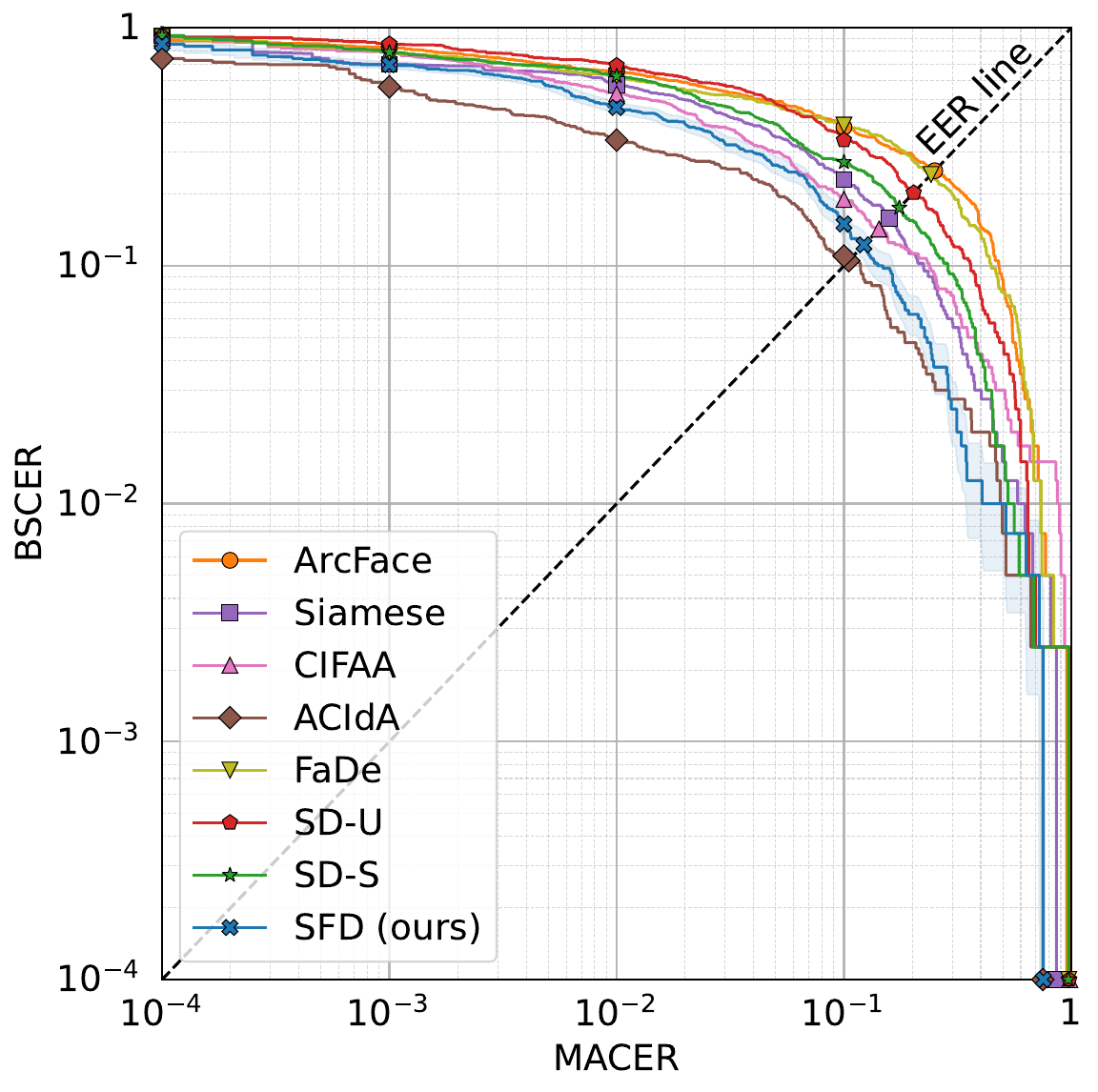}
        \caption{ArcFace FRS}
        \label{fig:ablation_arcface}
    \end{subfigure}
    \caption{Ablation studies on the FEI Morph V2~\cite{iciap2023, feimorph} dataset (criminal restoration scenario) presented as Detection Error Trade-off (DET) curves. (a) Impact of removing key training components (e.g., Feature Demorphing Module (FDM)). (b) Effect of varying the ratio of synthetic training data. (c) Robustness analysis against common image corruptions. (d) CurricularFace~\cite{curricularface} and (e) ArcFace~\cite{arcface} Face Recognition Systems (FRS) employed for the D-MAD decision-making of face demorphing methods.}
    \label{fig:ablation_dets}
\end{figure*}

Finally, to validate our design choices, we conduct ablation studies on FEI Morph V2~\cite{iciap2023, feimorph} under the challenging criminal restoration scenario, ensuring that the findings reflect the framework's performance in the most difficult setting.

\textbf{Impact of Training Components.}
Figure~\ref{fig:ablation_changes} analyzes the impact of various training components. We evaluate removing the Feature Demorphing Module (FDM), excluding UTW-NS~\cite{styledemorpher} or UTW-StyleGAN~\cite{styledemorpher} morphs from training, and omitting MS-SSIM loss. The results show that removing the FDM or excluding either morphing method leads to similar performance drops. While the MS-SSIM~\cite{ms-ssim} loss has a minimal individual contribution, the full configuration results in the lowest error rates. This confirms the FDM importance and demonstrates that exposure to both landmark-based and deep learning-based methods improves D-MAD.

\textbf{Synthetic Data Ratios.}
Figure~\ref{fig:ablation_syn_data} investigates the influence of synthetic training data. We vary the proportion of morphs derived from synthetic identities of FLUXSynID~\cite{fluxsynid} dataset from 0\% to 100\%, with the remainder composed of morphs generated using the real identities from DemorphDB~\cite{styledemorpher}. Notably, these corpora differ structurally: DemorphDB has fewer identities but offers a high volume of morphs per subject (up to 100), whereas FLUXSynID provides a vast number of identities with fewer morphs (up to 12).

The split comprising 80\% FLUXSynID data achieves the best D-MAD performance. Importantly, the model trained on 100\% synthetic data still outperforms all other D-MAD methods from Tab.~\ref{tab:dmad_results} by achieving an EER of 7.53\%. This suggests that effective demorphing methods can be trained predominantly on synthetic data, mitigating privacy concerns associated with using real biometric data during training.

\textbf{Robustness to Image Corruptions.}
Figure~\ref{fig:ablation_corruptions} evaluates robustness against common image degradations applied to both input images. We use the synthetic image corruption framework proposed in~\cite{corruptions} to apply brightness shifts, Gaussian noise, and JPEG compression at severity level 2. We simulate resolution loss by downsampling images to $128\times128$ pixels. We further simulate the print-and-scan process using the method proposed in~\cite{ubo_print_scan}, critical for evaluating physical handling artifacts. SFDemorpher maintains similar performance across all corruption types, with minimal deviation from the No~Corruption baseline. The print-and-scan simulation is most challenging, increasing EER from 5.33\% to 6.73\%. These results confirm SFDemorpher's generalizability and validate its suitability for real-world deployment.

\textbf{Impact of Face Recognition Systems.}
Figures~\ref{fig:ablation_curricularface} and \ref{fig:ablation_arcface} analyze the modularity of the proposed D-MAD pipeline (see Fig.~\ref{fig:dmad_setup}) by varying the underlying FRS. This change impacts all demorphing methods, as their decision logic relies on FRS scores. Traditional methods (Siamese, CIFAA, ACIdA) remain unaffected, operating independently of an external FRS. We evaluate CurricularFace~\cite{curricularface} and ArcFace~\cite{arcface} alongside the primary AdaFace~\cite{adaface} FRS previously shown in Fig.~\ref{fig:fei_crim}. 

As expected, performance correlates with FRS strength: AdaFace yields 5.33\% EER, CurricularFace 6.3\%, and ArcFace 12.3\%. With the weaker ArcFace FRS, traditional methods (e.g., ACIdA) outperform face demorphing approaches. However, two critical observations emerge. First, SFDemorpher outperforms all other face demorphing methods across all FRS models. Second, SFDemorpher improves upon the FRS baseline regardless of the FRS quality. This confirms the modular advantage of the demorphing pipeline that benefits from FRS technology advances. In contrast, traditional D-MAD methods do not rely on an external FRS, and those that incorporate FRS features typically require retraining.

\section{Conclusion} \label{sec:conclusion}
Face morphing attacks threaten travel document integrity~\cite{magic_passport, MAD, MAD_survey}, exploiting vulnerabilities from document issuance to verification. While traditional Differential Morphing Attack Detection (D-MAD) methods often rely on detecting low-level morphing artifacts, face demorphing shifts the paradigm to biometric consistency through identity disentanglement. In this paper, we introduced SFDemorpher, a framework leveraging joint StyleGAN~\cite{stylegan2} latent and feature space representations~\cite{sfe} to achieve high-fidelity reconstruction and robust D-MAD.

Extensive evaluation results demonstrate that SFDemorpher achieves state-of-the-art performance, outperforming existing traditional and demorphing-based D-MAD methods across multiple datasets. By training on both bona fide and morphed document images, our method yields superior distributional separability, evidenced by leading Bona Fide-Morph Separability (BMS) scores and balanced Deviation of Target/Non-Target Identity (DTI/DNTI) values. Furthermore, by generating reconstructed images rather than scalar scores, SFDemorpher provides visual explainability, aiding border guards and document officers in decision-making.

Crucially, SFDemorpher addresses generalizability, a primary barrier to operational deployment. Through a dual-pass training strategy and a predominantly synthetic corpus, the framework generalizes to unseen identities, diverse capture conditions, 13 morphing algorithms, and image corruptions including print-and-scan~\cite{ubo_print_scan} simulation. We extended the scope of evaluation to the challenging criminal restoration scenario, recently highlighted in~\cite{acida} to address enrollment-stage vulnerabilities beyond the traditional verification focus, where our method maintained SOTA performance despite the ill-posed nature of the task. The success of training predominantly on synthetic data further highlights the potential for scaling biometric security solutions while mitigating privacy concerns.

However, limitations remain. We observed performance variance between the FRLL-Morphs-UTW~\cite{frll_morphs_1, frll_morphs_2, styledemorpher} and FEI Morph V2~\cite{iciap2023, feimorph} datasets, particularly in distributional metrics. This likely stems from domain shifts, as synthetic FLUXSynID~\cite{fluxsynid} training data resembles the high-quality images from FRLL-Morphs-UTW, while the FEI Morph V2 introduces greater lighting and resolution variance. Although SFDemorpher demonstrates robust generalization, this gap indicates room for improvement in handling severe domain shifts.

\backmatter

\section*{Acknowledgments}
This research was funded by the European Union under the Horizon Europe programme, Grant Agreement No. 101121280. Views and opinions expressed are however those of the author(s) only and do not necessarily reflect the views of the EU/Executive Agency. Neither the EU nor the granting authority can be held responsible for them.

\begin{appendices}
\section{Architecture and Training Protocol} \label{appendix:implementation_details}
The pipeline in Fig.~\ref{fig:pipeline} comprises three trainable demorphing modules: $\mathcal{M}_{\text{IDM}}$, $\mathcal{M}_{\text{FDM}}$, and $\mathcal{M}_{\text{FFM}}$.

\textbf{Image Demorphing Module (}$\mathcal{M}_{\text{IDM}}$\textbf{).} 
We adopt the pre-trained Style-Feature Encoder~\cite{sfe} for this module. We modify the input convolutional layer by increasing channel dimension from 3 to 6 to accommodate the concatenated input pair $(I_{\text{doc}} \Vert I_{\text{ref}})$. The output is a feature tensor $F_{\text{IDM}}\in\mathcal{F}^9$.

\begin{figure}[htbp]
    \centering
    \includegraphics[width=0.60\linewidth]{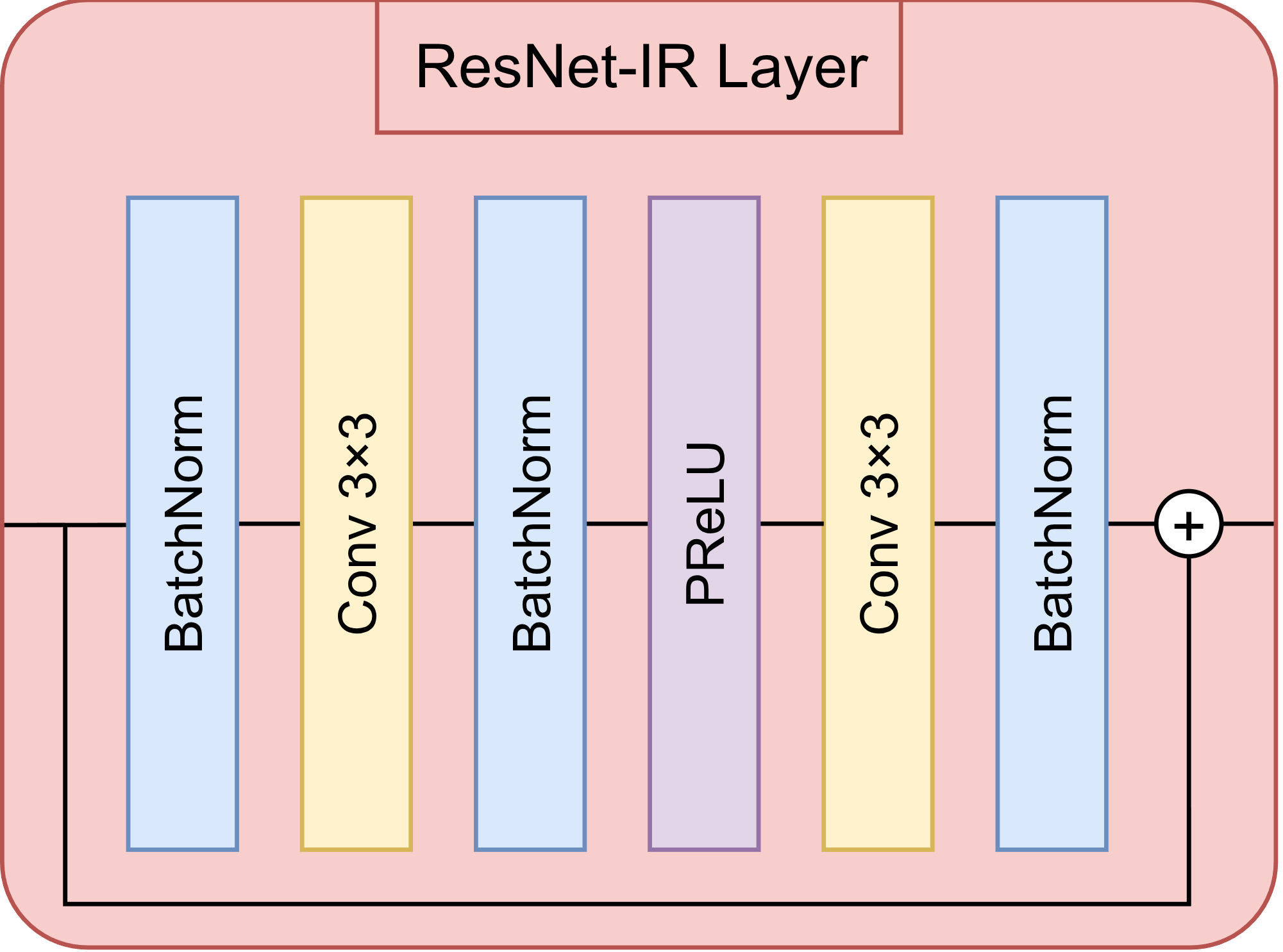}
    \caption{Structure of the ResNet-IR~\cite{arcface} layer.} 
    \label{fig:resnet_ir_layer}
\end{figure}

\textbf{Feature Demorphing and Fusion Modules (}$\mathcal{M}_{\text{FDM}}, \mathcal{M}_{\text{FFM}}$\textbf{).} 
Both modules consist of ResNet-IR~\cite{arcface} layers visualized in Fig.~\ref{fig:resnet_ir_layer}. 

\begin{figure}[htbp]
    \centering
    \includegraphics[width=0.85\linewidth]{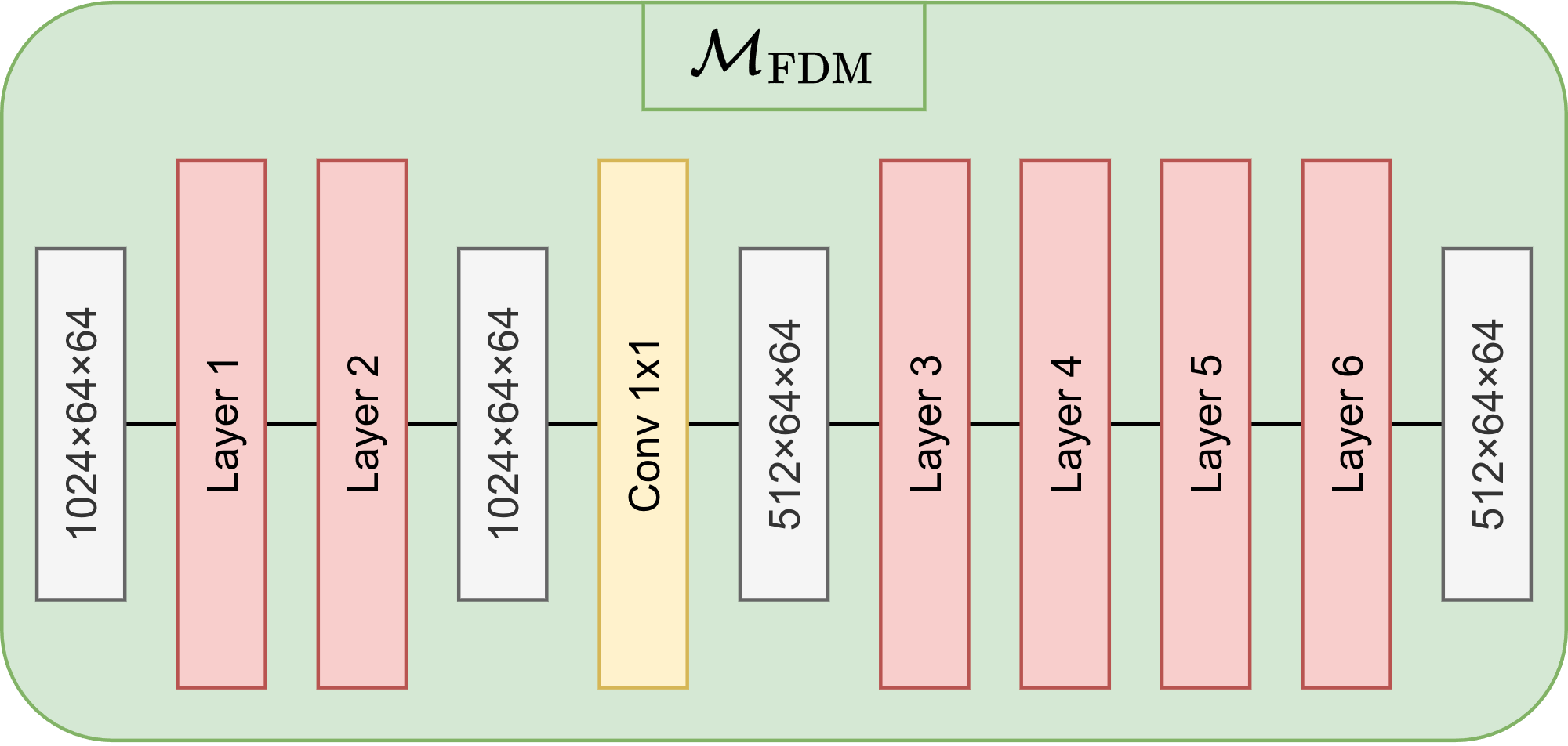}
    \caption{Architecture of the Feature Demorphing Module ($\mathcal{M}_{\text{FDM}}$).} 
    \label{fig:fdm}
\end{figure}

\begin{figure}[htbp]
    \centering
    \includegraphics[width=0.85\linewidth]{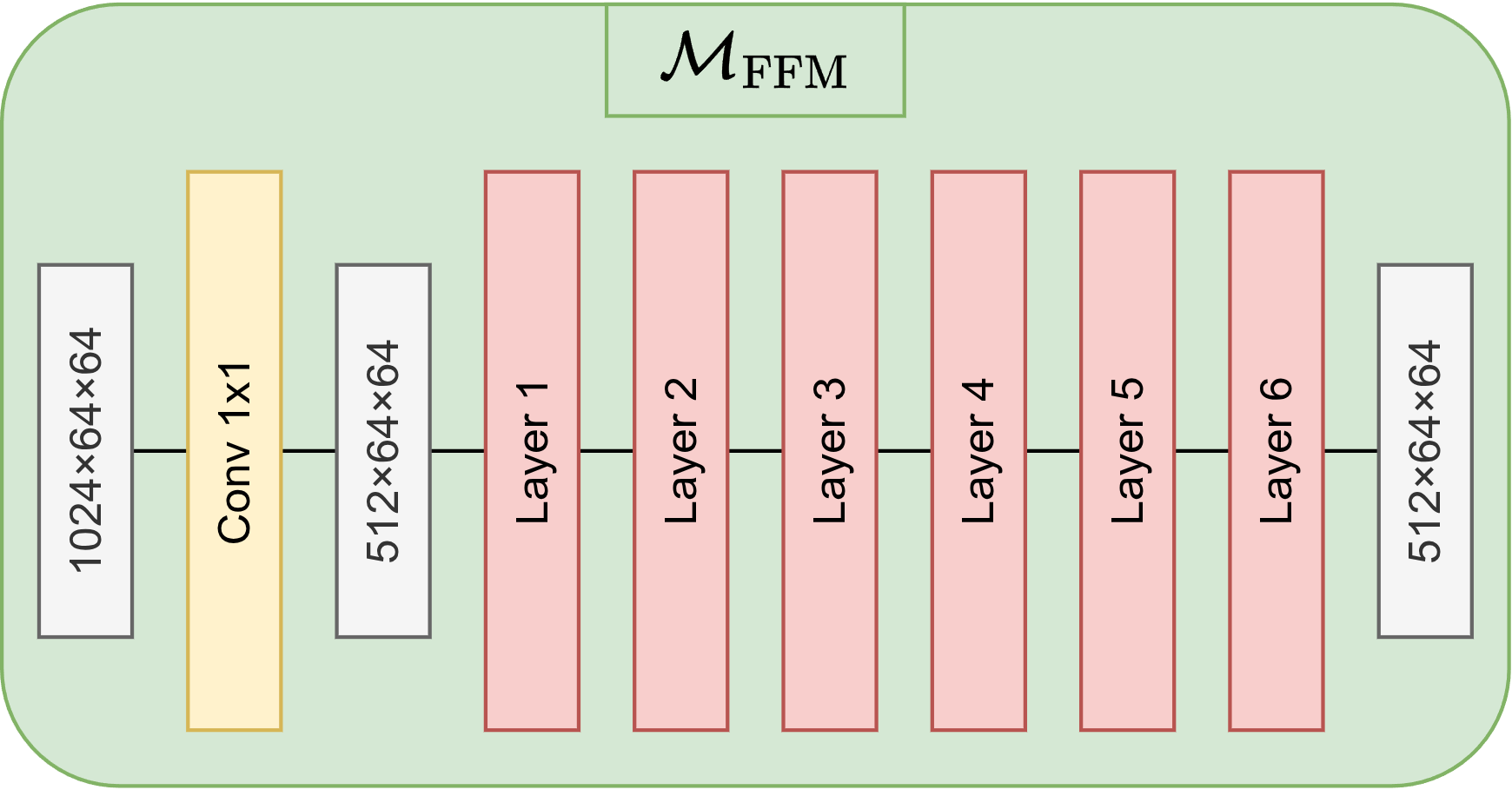}
    \caption{Architecture of the Feature Fusion Module ($\mathcal{M}_{\text{FFM}}$).} 
    \label{fig:ffm}
\end{figure}

\begin{itemize}
    \item $\mathcal{M}_{\text{FDM}}$ is trained from scratch using the architecture shown in Fig.~\ref{fig:fdm}. It processes concatenated features $(F_{\text{doc}} \Vert F_{\text{ref}})$ to perform demorphing in the feature space.
    \item $\mathcal{M}_{\text{FFM}}$ is initialized using the pre-trained weights of the Fuser network from the Style-Feature Encoder~\cite{sfe}. Its architecture is shown in Fig.~\ref{fig:ffm}. It fuses the outputs of $\mathcal{M}_{\text{FDM}}$ and $\mathcal{M}_{\text{IDM}}$ modules $(F_{\text{FDM}} \Vert F_{\text{IDM}})$.
\end{itemize}

\begin{table*}[htbp]
\caption{Morphing Attack Potential (MAP)~\cite{map_metric} scores for each morphing method across evaluation datasets. Values represent the percentage of morphs successfully verifying against both contributing subjects on at least $c$ FRSs (with $r=1$). Higher values indicate greater attack potential.}
\label{tab:map_results}
\centering
\footnotesize
\begin{tabular*}{\textwidth}{@{\extracolsep\fill}ccccc}
\toprule
\textbf{Dataset} & \textbf{Morphing Method} & \multicolumn{3}{c}{\textbf{\# of FRSs ($c$)}} \\
\cmidrule{3-5}
 & & \textbf{1} & \textbf{2} & \textbf{3} \\
\midrule
\multirow{5}{*}{FRLL-Morphs-UTW}
& UTW & 76.8\%  &  57.6\%  &  20.2\%  \\
& UTW-StyleGAN & 81.7\% &  63.4\%  & 41.6\% \\
& OpenCV &  95.9\% &  90.1\%  &  66.3\%   \\
& FaceMorpher & 94.4\% &   87.6\%  &  61.8\%  \\
& WebMorph & 96.2\% & 89.9\% &  65.1\%  \\
& AMSL & 91.0\% &  80.1\% &  44.4\%  \\
\midrule
\multirow{1}{*}{HNU-FM} 
& Landmark-Based & 98.5\% & 94.7\% & 85.3\%\\
\midrule
\multirow{6}{*}{FEI Morph V2} 
& C02 & 97.6\% & 93.8\% & 78.2\%  \\
& C03 & 76.8\% & 64.5\% & 46.0\% \\
& C05 &  84.1\% & 72.1\% & 54.4\%\\
& C08 & 85.2\% &  76.1\%  & 56.2\%  \\
& C15 & 80.2\% & 67.3\% & 46.3\%  \\
& C16 & 91.1\% & 84.1\%  & 62.2\%  \\
\bottomrule
\end{tabular*}
\end{table*}

Both modules receive concatenated feature maps of size $1024\times64\times64$ (derived from two $512$-channel inputs) and output a feature tensor of size $512\times64\times64$ corresponding to $\mathcal{F}^9$.

We jointly train these modules ($\mathcal{M}_{\text{FDM}}, \mathcal{M}_{\text{IDM}}, \mathcal{M}_{\text{FFM}}$) using the Ranger optimizer (Lookahead technique~\cite{lookahead} combined with the Rectified Adam~\cite{rect_adam} optimizer) with a learning rate of $5 \times 10^{-5}$. The pre-trained Discriminator~\cite{stylegan2, sfe} is optimized via Adam optimizer~\cite{adam} with a learning rate of $1 \times 10^{-4}$. The framework is trained for 68,000 iterations with a batch size of 2.

To stabilize optimization during the morphed pass, we employ a curriculum-based sampling strategy for the trusted reference image $I_{\text{ref}}$. Initially, to minimize domain variance, $I_{\text{ref}}$ is selected as the exact constituent document image used to generate the morph. The scheduler then linearly increases the probability of instead sampling $I_{\text{ref}}$ from the challenging live capture domain, capping at a maximum of 80\% at step 40,000. This ensures the model first converges on the fundamental demorphing task before adapting to the cross-domain variations of trusted reference images.

\section{Morphing Attack Potential} \label{appendix:map}
We assess the inherent risk posed by each morphing technique using the Morphing Attack Potential (MAP) metric~\cite{map_metric}. MAP quantifies the attack potential by measuring the proportion of morphed images that successfully verify against both contributing subjects across multiple Face Recognition Systems (FRSs) and verification attempts.

The original MAP definition~\cite{map_metric} constructs a matrix MAP$[r, c]$, where $r$ represents the number of verification attempts (probe images) and $c$ represents the number of FRSs. For a given dataset, each matrix entry represents the percentage of morphed images that achieve successful verification against both contributing identities across a minimum of $r$ probe samples and $c$ distinct FRSs. A higher MAP value indicates a higher attack potential, meaning the morphing method generates images capable of fooling multiple FRSs simultaneously.

While the original MAP formulation accounts for multiple probe images per subject (e.g., video frames at a border gate), our evaluation protocol utilizes a single probe image per subject ($r=1$). Consequently, we solely focus on the generality dimension ($c$).

Table~\ref{tab:map_results} reports the MAP scores for each morphing method across the evaluation datasets. We evaluate against three distinct FRS backbones (AdaFace~\cite{adaface}, CurricularFace~\cite{curricularface}, and ArcFace~\cite{arcface}). The decision threshold of each FRS was set to yield $0.01\%$ False Match Rate (FMR) on DemorphDB~\cite{styledemorpher} dataset.

The results in Tab.~\ref{tab:map_results} reveal notable differences in attack potential across morphing techniques. OpenCV~\cite{opencv_morphs}, FaceMorpher~\cite{facemorpher}, WebMorph~\cite{webmorph}, and HNU-FM~\cite{hnu_fm} morphing methods exhibit the highest MAP scores, with HNU-FM reaching 85.3\% for $c=3$. However, these methods generate full-face morphs without splicing~\cite{splicing} with visible ghosting artifacts, making them less likely to pass human inspection despite fooling multiple FRSs.

In contrast, splicing-based methods (UTW~\cite{utw_morphs}, AMSL~\cite{amsl}, FEI Morph V2~\cite{iciap2023, feimorph} morphs) show lower MAP scores, with UTW at only 20.2\% for $c=3$. These methods embed the morphed inner face into one of the contributor's outer facial structure, creating visually realistic morphs harder for humans to detect. Notably, the deep learning-based UTW-StyleGAN~\cite{styledemorpher} morphs achieve higher attack potential (41.6\% for $c=3$) by generating full morphs without ghosting artifacts, challenging both FRSs and human inspectors.

These results highlight that high FRS vulnerability does not necessarily correlate with real-world attack success. Methods with lower MAP scores but higher visual fidelity can represent the more pressing security concern, as they can bypass both automated systems and human inspectors.
\end{appendices}

\bibliography{references}

\end{document}